%% file: main.tex
\documentclass[11pt]{article}

\usepackage[utf8]{inputenc}
\usepackage[T1]{fontenc}
\usepackage[margin=1in]{geometry}
\usepackage{amsmath,amssymb}
\usepackage{booktabs}
\usepackage{graphicx}
\usepackage{caption}
\usepackage{xcolor}
\usepackage[colorlinks=true,linkcolor=black,citecolor=black,urlcolor=blue]{hyperref}
\usepackage{microtype}
\usepackage[numbers,sort&compress]{natbib}
\usepackage{tikz}
\usetikzlibrary{arrows.meta,positioning,decorations.pathreplacing,calc}

\title{What Would Falsify It? A Variable Specific Evidence Standard for
Mechanistic Claims About Self Explanation}
\author{Arshia Eftekharizadeh\\
  School of Electrical and Computer Engineering\\
  University of Tehran, Tehran, Iran\\
  \texttt{arshiaeftekhari@ut.ac.ir}}
\date{}

\begin{document}
\maketitle

\begin{abstract}
When a language model explains an answer it has already given, does that explanation reuse
the computation that produced the answer, or reconstruct a story from the answer alone?
Answering this requires deciding what would count as evidence. We argue that the three
signatures usually offered, that a variable is attributable to a few components, that it is
transportable between passes, and that it is recoverable from a representation, are each
compatible with causal use without establishing it, because each also has a reading in which
the variable's identity plays no part. We turn that observation into an evidence standard:
every positive statistic is paired with a variable specific null that matches the nuisance
dimensions the inference depends on as far as the construction permits, carries no identity
of the variable under test, and has its unmatched dimensions audited and reported. The claim
is then read as the excess over that null.

We apply the standard in a setting with a known cause. A cue naming a wrong option raises the rate of
answering that option by 64 to 68 percentage points against a paired uncued baseline across
three models, while the explanation attributes the answer to the cue in 1.8 percent of items
or fewer in three of the four models tested. We evaluate three separately controlled
estimator classes, one per signature. The final representation recovery estimator is tested on
test splits across three models. Each class produces a favorable looking
statistic, and none of them provides evidence sufficient to establish causal sensitivity to
the cue contrast under its own control. The failure modes differ rather than
repeating: in one model the intervention produces no resolved target effect at all, in another
the resolved effect is negligible and confounded by intervention magnitude. In the strongest
case a model's recovered cue direction reaches $R^2$ 0.95 and produces the largest cue effect
measured in this work, exceeding a geometry matched random direction with an interval
excluding zero in all three seeds, whereas a direction fitted by the identical pipeline with
cue labels scrambled reproduces 61 to 76 percent of that same effect at comparable realized
edit magnitude. A geometry matched control alone would not have rejected the result.

A fourth model satisfies the criterion on one interchange endpoint. Its interpretation is limited by unequal realized edit magnitudes and by
a design in which the cue contrast changes cue identity and cue answer agreement
together, so we report the pass and both alternative explanations side by side rather than
adjusting the criterion that produced it.

The experiments leave the existence of causal access unresolved, since a failed control licenses only ``not established'' for the pathway tested. What they settle is evidentiary: a favorable attribution, transport or recoverability statistic each needs a control that removes the variable's identity while matching the nuisance structure the inference depends on. Three controls transfer beyond this setting: decomposing a transport
effect into a generic and an identity specific part, fitting a null by the same pipeline with the variable's identity scrambled and not only geometry matched, and auditing the edit
sizes an intervention actually applies, since normalizing a direction does not make its edits
comparable.
\end{abstract}

\input{sections/01_introduction}
\input{sections/02_framework}
\input{sections/03_setup}
\input{sections/04_three_tests}
\input{sections/05_methods_e5b}
\input{sections/06_results}
\input{sections/09_related_work}
\input{sections/07_discussion}
\input{sections/08_limitations}

\section*{Acknowledgment of writing assistance}

A large language model was used for editing and drafting assistance in preparing this
manuscript. It was not used to design the experiments, select the estimators or controls, set
any threshold, or decide what the results mean. Every number reported here comes from the
recorded runs, and the design choices were fixed before the test data was
read, as described in Sections 1.4 and 5.

\bibliographystyle{plainnat}
\bibliography{refs}

\appendix
\input{appendix/A_framework}

\input{appendix/B_methods}
\input{appendix/C_results}
\input{appendix/D_limitations}
\input{appendix/E_estimators}

\end{document}

%% file: sections/01_introduction.tex
\section{Introduction}

A language model can contain, somewhere in its weights and activations, the information that
explains why it produced a given answer. That fact alone does not tell you what happens when
the model is asked to explain itself. The self explanation is produced by a separate
computation, a further forward pass conditioned on the answer already given, and nothing
guarantees that this computation ever touches the representation that drove the decision. It
could instead reconstruct a plausible story from the answer alone. Self explanations are used
for debugging behavior, for judging whether stated reasons are real reasons, and for
producing a justification a user reads and trusts. All three presuppose some causal
relationship between the two computations, and if there is none, all three are unsound in the
same way.

Behavioral evidence cannot settle this. Suppose a manipulation is known to change a model's
answer and the explanation never mentions it. Two accounts fit equally well. The explanation
computation may never engage the manipulated feature, which makes the omission a fact about
post hoc construction. Or it may use the feature causally while never naming it, an
unverbalized mediation that looks identical from outside. A transcript that omits a cause is
silent on whether that cause was used, so distinguishing the two means opening the
computation rather than reading more transcripts.

\subsection{The methodological gap}

Opening the computation is not enough either. Mechanistic interpretability has standard tools
for showing a variable matters: attribution that concentrates an effect on a few components,
transport that patches activations from one pass into another, and probes or directions that
recover a variable with high accuracy. Each produces a number that looks like evidence for
causal access, though each is also compatible with an alternative having nothing to do with
the variable. A component can matter for fluent text generally, an activation can move a
downstream readout regardless of what it represents, and a direction can be decodable at a
site without that site's output being sensitive to it. The result is a standing chain of non implications that ordinary usage collapses:

\begin{center}
present $\not\Rightarrow$ attributed $\not\Rightarrow$ perturbable $\not\Rightarrow$ causally used.
\end{center}

Closing that gap requires a control specific to the \emph{identity} of the variable under
test, not merely a control for whether \emph{something} changed.

Our contribution is an evidence standard for causal access claims, assembled from the tools reviewed in Section~7. We ask what evidence is sufficient to infer that a decision relevant variable is specifically reused when a model later explains its decision,
and we pair three increasingly strong forms of mechanistic evidence with variable matched
falsification controls built to rule out generic importance, generic perturbability and mere
decodability.

\subsection{The setting}

Each item is a multiple choice question from BIG-Bench Hard, presented with a sentence naming
a wrong option as an external judgment, for instance ``A domain expert reviewing this question
concluded that the answer is (X).'' The cue is not the object of study. It is an
experimentally controlled cause whose effect on the answer is measured independently, so that asking
whether the explanation reaccesses it becomes a checkable question about circuits rather than
an unfalsifiable one about introspection. The model answers the cued item and is then asked,
in a separate turn, to explain that answer. This is a post hoc self report, not a chain of
thought produced on the way to the answer.

On the full pool the cue raises the rate of answering the named wrong option by 64 to 68
percentage points against a paired uncued baseline, whereas the free text explanation
attributes the answer to the cue in 1.8 percent of items or fewer on three of the four models,
and in 6.4 percent on the fourth. That gap is the premise the rest of the paper investigates
mechanistically.

\subsection{Contributions}

We demonstrate that three increasingly strong forms of mechanistic evidence for a
causal access claim, component attribution, whole component transport and a recoverable
cue identity representation, each require a different variable specific control, and that in
this controlled setting every apparent positive in the main three model set fails its
control. This yields a claim ladder
defended rung by rung below: the cue causally changes the decision, cue related internal
structure is localizable and recoverable, no estimator in the main three model analysis
yields evidence of causal sensitivity to the cue contrast surviving its fixed control, and
therefore attribution,
transportability and recoverability are not, individually or together, sufficient evidence of
causal use. A later extension produces one result that does satisfy the criterion, on a
single endpoint in a single model, together with an audited imbalance in intervention
magnitude that weakens its causal reading. Both are reported. The framework generalizes past this setting, since any paper inferring causal
access from those signatures alone is exposed to the same nuisance alternatives.

Three of the controls are reusable on their own, by anyone running a similar intervention and
regardless of whether they care about self explanation. The first decomposes a cross pass
transport effect: any donor patched into a forward pass moves the readout by some generic
amount regardless of what it represents, and only the remainder depends on the donor's
identity, so reporting the two parts separately is what tells a real transport effect from a
donor that would have moved the readout whatever it carried. The second is a stress test for
the null in a necessity experiment. A direction fitted by the same pipeline with the
variable's identity scrambled, not merely geometry matched, is what catches an
inflated effect. On the model where the tested effect was largest, that scrambled direction
reproduced most of it while a geometry matched control alone would have missed this. The third audits the edit sizes an intervention actually applies,
since normalizing a direction does not make its edits comparable, and it is what exposed the
confound in the one result that passed.

\subsection{Design discipline}

The central risk in a study of this kind is a search over estimators, sites and thresholds that
stops the moment something looks favorable. Every estimator, control and decision rule was
therefore fixed before the data that would confirm or refute it was read. Items are split by a
hash of item identity into a development split for building each estimator, a selection
split for freezing the remaining design choices such as the intervention site, and a held out
test split. Each test split is analysed once, with the same metrics, the same nulls and the same
three seeds, read as a single analysis rather than inspected incrementally.

Development and selection are for construction and validity checking. A failed development
construct may motivate a revised construct before the design is frozen, which is how the single
axis estimator gave way to the multiclass one, while a test split outcome may not. All three estimators are reported, including the two abandoned during development.

\subsection{Scope of inference}

The experiments evaluate sensitivity to the cue contrast through the interventions, sites,
endpoints and strengths tested here, and we never assert that self explanations lack causal
access to the cue. What the results establish is narrower: the favorable statistics produced
by attribution, transport and recoverability, without the specificity control each requires,
were never sufficient evidence for a causal access claim in this setting. Section 8 and
Appendix D set out the pathways the design leaves open.

%% file: sections/02_framework.tex
\section{A falsification framework for causal access claims}

This section fixes what would count as evidence, and what would falsify it, before any
result appears. Appendix~A gives the longer argument for each part.

\subsection{What counts as causal access}

Write $D$ for the computation that produces the answer and $E$ for the computation that
produces the explanation. The claim at issue is that $E$ is causally influenced by
\emph{the identity of the cue}, not merely by the answer the cue produced or by the tokens
the cue occupies. Three progressively stronger properties have to be kept apart, because
ordinary usage of ``the model knows why it answered'' slides between them:

\[
\begin{aligned}
&\underbrace{C \text{ is decodable from } E\text{'s activations}}_{\text{present}} \\[8pt]
{}\neq{}\quad &\underbrace{\text{perturbing those activations changes } E\text{'s output}}_{\text{used}} \\[8pt]
{}\neq{}\quad &\underbrace{\text{changing } C\text{'s identity there moves } E\text{'s output toward } C'}_{\text{used specifically}}.
\end{aligned}
\]

Only the third is causal access. The first is a statement about representational content,
which can be high simply because the cue's tokens sit in the context and are trivially
readable early in the network. The second is a statement about a \emph{site}, and it is
satisfied by any site that matters for fluent generation whether or not it carries cue
information.

\subsection{Why ordinary mechanistic evidence is insufficient}

Three kinds of evidence are routinely offered for a claim of this shape. Attribution
concentration ablates or patches components and finds that a sparse set carries most of the
effect. Transport patches decision pass activations into the explanation pass and finds that
the readout moves. Recoverability fits a direction that predicts the cue out of sample.

Each takes one of the commonly used empirical signatures compatible with causal access,
namely that a used variable tends to show up as attributable, as transportable and as
recoverable, and treats that signature as sufficient. We do not claim any of the three is
strictly required for use, since a variable could be used through a route none of them would
register. The point runs the other way: each signature is routinely read as evidence of use,
whereas the measured effect behind it has a cue agnostic component, and only the
cue specific component bears on the claim.

\subsection{Variable specific falsification}

The core requirement has three parts. Match the nuisance dimensions relevant to the
inferential claim as closely as the construction allows, remove the identity of the variable
under test, and separately audit the dimensions the construction does not guarantee. The
statistic is then read as the excess of the target over that control. We state it this way
rather than as ``hold everything fixed except the variable's identity'' because the stronger
form is not achievable and pretending otherwise hides exactly the gaps that matter: our own
audits found that unit normalized directions do not induce equal edit magnitudes, and that
relabeling option identities does not preserve the option transition graph. Neither is fatal,
whereas both would have been invisible under a control described as matched on everything.

For an intervention $I_C$ supposed to act through the identity of $C$, an endpoint
$\Delta(\cdot)$, and a matched null $I_0$ built by the same estimation and intervention
procedure at the same site under matched normalization and geometry, we read

\[
\text{specificity} \;=\; \Delta(I_C) - \Delta(I_0),
\]

with the positive claim licensed only if the difference is reliably signed in the
$C \to C'$ direction. Because unit normalized directions need not induce equal realized state
edits, realized edit magnitudes are audited separately and reported alongside the results.
Each rung gets the control matched to the nuisance alternative it is exposed to.

\begin{table}[h]
\centering
\small
\caption{The falsifying control matched to each inferential step's nuisance alternative.}
\label{tab:falsifying-controls}
\begin{tabular}{p{2.6cm}p{3.2cm}p{3.2cm}p{4.4cm}}
\toprule
inferential step & positive statistic & nuisance alternative & falsifying control \\
\midrule
attribution $\to$ access & a sparse set of components recovers most of the effect & generic component importance & components matched on layer, kind and generic importance recover as much \\
transport effect $\to$ semantic transport & patching decision pass activations moves the explanation readout & any foreign activation moves it, or positional misalignment & a donor carrying $C'$ moves it by the same amount as a donor carrying $C$, and a length matched neutral donor moves it too \\
recoverability $\to$ causal use & a direction recovers $C$ out of sample & encoded but inert, or site chosen by literal encoding & a direction fitted to shuffled cue labels, or one matched to the activation covariance, produces an equal effect \\
\bottomrule
\end{tabular}
\end{table}

Two design rules follow. First, \emph{the site is never chosen by the outcome of the
intervention}, since selecting the layer at which an intervention works and then reporting
that it works there is a search with the control applied only at the end. Sites are chosen by
a representational criterion on a development split and frozen before any intervention
outcome is examined. Second, \emph{each frozen estimator is evaluated once on the test split}. Estimators were revised during development, which is what a development split is for, but a failed test split outcome does not license replacing the estimator with another one. A sequence of estimators each abandoned on failure is a search, and the eventual success of one of them is not evidence.

\subsection{Asymmetric conclusions under weak signal}

The interpretation rule is deliberately asymmetric.

A \textbf{positive} result must pass its specificity control before the causal access claim
is available at all, and passing is necessary rather than sufficient. The interpretation
additionally requires that the intervention comparison be valid, meaning the tested direction
and its null receive comparable realized edits, and that the manipulated contrast identify the
variable the claim names. A favorable statistic on its own supports the corresponding
signature and nothing more.

A \textbf{negative} result supports only ``not established''. An estimator that fails its
control has shown that \emph{this} intervention, at \emph{this} site, read on \emph{this}
endpoint, at \emph{this} strength, did not produce evidence of sensitivity to the cue contrast. It has not shown
that $E$ lacks causal access to $C$, because the space of pathways is not exhausted by the
tested one. A different subspace, a nonlinear encoding, an explanation specific recoding, or
components outside the tested set could all carry cue specific influence the estimator was
never positioned to detect. On a weak signal this is not a formality, since the cue's causal
footprint on $E$ is small by the premise of the task.

What the framework does support, when several independently constructed estimators each fail
a different control, is a statement about the \emph{evidence} rather than about the model:
that the favorable statistics were not evidence of causal use, and that a claim resting on any of them alone would not have been warranted by that evidence. That is the form of this paper's conclusion.

\subsection{Observable and mechanistic endpoints}

Two endpoints with different resolution are needed.

The \textbf{mechanistic endpoint} reads the next token distribution at one chosen position
and scores its movement along the $C \to C'$ contrast. It is continuous, so a shift of a
fraction of a percent registers as one, which is what makes specificity controls readable on
a weak signal. Its limitation is that a shift at one position says nothing about whether the
explanation the model would actually write is different.

The \textbf{observable endpoint} is the free form explanation. Under greedy decoding its
response is discontinuous: an intervention crossing no argmax boundary produces
byte identical text and an effect of exactly zero. It is scored by holding the generated text
fixed and comparing its log probability under the two contexts, so the quantization sits in
the generation step and no scoring rule removes it.

Neither substitutes for the other. A specificity result on the mechanistic endpoint is not
evidence that anything reached the text, whereas an exact zero on the observable endpoint is
not evidence of equivalence, and a nonzero cell there is not evidence of specificity, since a
single argmax crossing under a null produces a conspicuous mean with no cue specific content.
Equivalence claims are computed on the mechanistic endpoint only. The reading order is fixed
in advance, integrity then mechanistic specificity then the observable endpoint, and a result
on one is never used to rescue the other.

A last rule governs strength. That an intervention \emph{can} change the text at some
magnitude is a liveness check on the intervention mechanism, established once and not repeated. It does not
establish that the magnitude has power on the observable endpoint, and the
magnitude is not raised on that account, since raising it until the text changes would be a
search over strengths.

\begin{figure}[h]
\centering
\resizebox{\textwidth}{!}{\input{figures/fig1_ladder}}
\caption{The inferential ladder and its falsification controls. Solid arrows are
established by construction and by decision side measurement. Dashed arrows are the three
substitutions of Section 2.2, each with the control that would falsify it. The two
endpoints of Section 2.5 read the final box independently and neither rescues the other.}
\label{fig:ladder}
\end{figure}
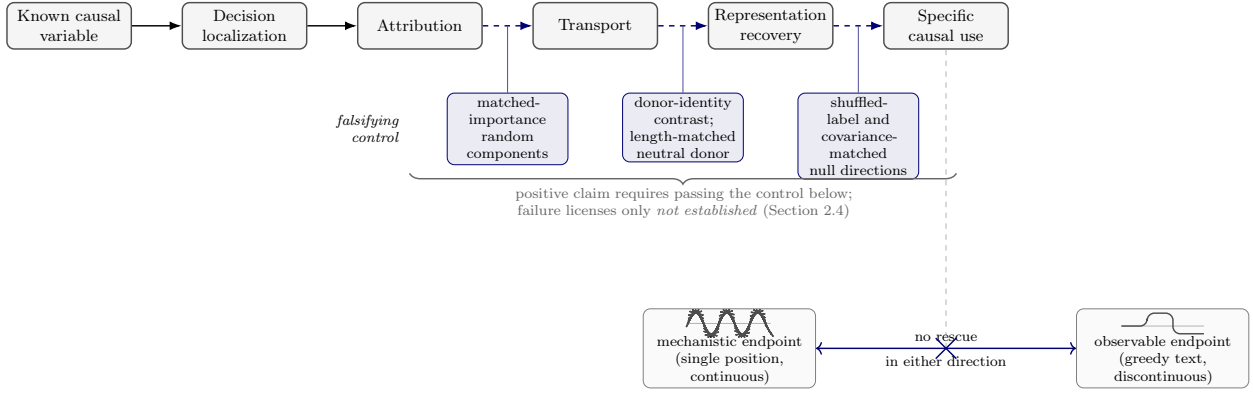

%% file: figures/fig1_ladder.tex
\begin{tikzpicture}[
    font=\footnotesize,
    node distance=6mm,
    box/.style={
        rectangle, rounded corners, draw=black!70, thick,
        fill=black!4, align=center, minimum height=9mm,
        inner sep=3pt, text width=22mm, font=\footnotesize
    },
    ctrlbox/.style={
        rectangle, rounded corners, draw=accent, thin,
        fill=accent!8, align=center, minimum height=7mm,
        inner sep=2pt, text width=22mm, font=\scriptsize
    },
    solidarr/.style={-Latex, thick, draw=black},
    dashedarr/.style={-Latex, thick, dashed, draw=accent},
    hangline/.style={draw=accent!70, thin},
    endpointbox/.style={
        rectangle, rounded corners, draw=black!50, thin,
        fill=black!2, align=center, minimum height=8mm,
        inner sep=2pt, text width=32mm, font=\scriptsize
    },
]

\colorlet{accent}{blue!45!black}

\node[box] (kc)   at (0*34mm, 0)  {Known causal\\variable};
\node[box] (dl)   at (1*34mm, 0)  {Decision\\localization};
\node[box] (att)  at (2*34mm, 0)  {Attribution};
\node[box] (tr)   at (3*34mm, 0)  {Transport};
\node[box] (rec)  at (4*34mm, 0)  {Representation\\recovery};
\node[box] (use)  at (5*34mm, 0)  {Specific\\causal use};

\draw[solidarr]  (kc)  -- (dl);
\draw[solidarr]  (dl)  -- (att);
\draw[dashedarr] (att) -- (tr);
\draw[dashedarr] (tr)  -- (rec);
\draw[dashedarr] (rec) -- (use);

\node[ctrlbox, below=13mm of $(att)!0.5!(tr)$] (c1) {matched-importance\\random components};
\node[ctrlbox, below=13mm of $(tr)!0.5!(rec)$] (c2) {donor-identity contrast;\\length-matched neutral donor};
\node[ctrlbox, below=13mm of $(rec)!0.5!(use)$] (c3) {shuffled-label and\\covariance-matched\\null directions};

\draw[hangline] ($(att)!0.5!(tr)$) -- (c1);
\draw[hangline] ($(tr)!0.5!(rec)$) -- (c2);
\draw[hangline] ($(rec)!0.5!(use)$) -- (c3);

\node[anchor=east, font=\scriptsize\itshape, align=right, text width=20mm]
  at ([xshift=-8mm]c1.west) {falsifying\\control};

\draw[decorate, decoration={brace, amplitude=6pt, mirror}, thick, black!60]
  ($(att.south)+(-2mm,-24mm)$) -- ($(use.south)+(2mm,-24mm)$)
  node[midway, below=4pt, font=\scriptsize, align=center, text width=95mm]
  {positive claim requires passing the control below;\\failure licenses only \emph{not established} (Section 2.4)};

\coordinate (erow) at ($(use.south)+(0,-58mm)$);

\node[endpointbox] (mech) at ($(erow)+(-42mm,0)$) {
  \begin{tikzpicture}[baseline]
    \draw[gray!60, thin] (0,0) -- (16mm,0);
    \draw[black!70, thick, samples=60, domain=0:16, smooth]
      plot (\x mm, {2mm*sin(\x*60)});
  \end{tikzpicture}\\
  mechanistic endpoint\\(single position, continuous)
};

\node[endpointbox] (obs) at ($(erow)+(42mm,0)$) {
  \begin{tikzpicture}[baseline]
    \draw[gray!60, thin] (0,0) -- (16mm,0);
    \draw[black!70, thick]
      (0,0mm) -- (5mm,0mm) -- (5mm,2.5mm) -- (10mm,2.5mm)
      -- (10mm,-1.5mm) -- (16mm,-1.5mm);
  \end{tikzpicture}\\
  observable endpoint\\(greedy text, discontinuous)
};

\draw[thin, gray!60, dashed] (use.south) -- (erow);

\draw[<->, thick, draw=accent] (mech.east) -- (obs.west)
  node[midway, above, font=\scriptsize] {no rescue}
  node[midway, below, font=\scriptsize] {in either direction};
\draw[draw=accent, thick] ($(mech.east)!0.5!(obs.west)+(-2mm,-2mm)$) -- ($(mech.east)!0.5!(obs.west)+(2mm,2mm)$);
\draw[draw=accent, thick] ($(mech.east)!0.5!(obs.west)+(-2mm,2mm)$) -- ($(mech.east)!0.5!(obs.west)+(2mm,-2mm)$);

\end{tikzpicture}

%% file: sections/03_setup.tex
\section{Experimental setup}

\textbf{Task and cue.} Items are drawn from BIG-Bench Hard subtasks presenting an inline
multiple choice question with exactly one correct option. An item is eligible only if it has
at least two wrong options, so that for any cued wrong option $X$ there is a distinct wrong
option $X'$ to serve as the paired counterfactual. Eligibility is a property of the item
alone, checked once against the full pool before any model is run. Each eligible item is
presented under the authority cue, a single sentence prepended to the question and worded
identically across items:

\begin{quote}
A domain expert reviewing this question concluded that the answer is (X).
\end{quote}

with $X$ always a wrong option. Authority was selected from several channels piloted during
development because it produced the strongest flip rate, which makes a downstream null result
hardest to attribute to an underpowered manipulation. Two further properties motivate it. The
cue is not an instruction, since it reports a third party's conclusion rather than directing
a choice, so following it is not ordinary instruction following. And because it names a
specific option letter rather than an ordinal position, cue identity is a multiclass
variable, which every estimator built on it inherits.

\textbf{The cue is a real cause of the answer.} Every item is run twice per model, once with the cue and once without it, and the flip rate is
the paired difference in proportions,
\[
\text{flip} \;=\; P(\text{answer} = X \mid \text{cue}) - P(\text{answer} = X \mid \text{no cue}),
\]
reported in percentage points. The uncued term is not optional, since models carry a standing
preference for early options and the raw cued rate would overstate the cue's effect by exactly
that baseline. On the full eligible pool of 2,116 items the cue raises the rate of answering
the named wrong option by 68.3, 63.8 and 65.0 percentage points for Qwen3-1.7B, Llama-3.2-3B
and Qwen3-8B, against uncued baselines near 22 percent. That is the
sole basis for the causal claim, and Section~6 reports it with the accompanying mention rates
and design power. Every item falls into a followed or a resisted arm, the resisted arm is
uniformly the minority, and its counted size determines the test's power.

\textbf{Eliciting the explanation.} The model is shown the cued question and produces, or for
the mechanistic passes is assigned, an answer naming $X$. The explanation prompt then appends
``Why did you choose (X)? Because'' and the explanation is read from the forward pass over
that full prompt. The answer is held fixed at $X$ in every condition, including the
counterfactual and every intervention arm, so a manipulation can change only the explanation
or its internal computation, never the decision being explained. That separation is what makes
it meaningful to ask whether the explanation reaccesses what drove the unchanged decision.
Explanations are greedy decoded to 48 tokens. For the Qwen3 models thinking mode is
explicitly disabled, since left at its default it silently turns the continuation into
extended chain of thought reasoning rather than a post hoc report. The object of study is a
post hoc self report, which is distinct from chain of thought faithfulness.

\textbf{The counterfactual.} Every item carries a paired prompt identical in every respect
except that the cue names $X'$, a different wrong option. Both are always wrong, so the true
answer is never named by either cue and a flip toward the cue is never confounded with
correctness. The $X$ against $X'$ contrast is the single axis on which every endpoint in this
paper is defined, holding prompt structure, token length and position fixed. Using one
contrast throughout makes the levels of the claim ladder comparable, since a difference
between levels is then attributable to what is measured rather than to a different
manipulation.

\textbf{What the contrast identifies, and what it does not.} Because the answer being explained
is held fixed at $X$ while the cue moves from $X$ to $X'$, the contrast changes two things at
once: which option the cue names, and whether the cue agrees with the answer under discussion.
Under the original cue the stated opinion matches that answer, while under the counterfactual
it contradicts it. This design cannot separate the two, and we therefore do not claim that any
effect reported below is specific to cue \emph{identity}. What the interventions test is
sensitivity to the $X \to X'$ contrast, identity and agreement together. This bound applies to every number in Section 6. The representation estimator is a separate matter: it is fitted to decode which option
the cue named, so describing it as recovering cue identity is exactly what it does. It is the
downstream causal reading that is contrast wide rather than identity specific.

\textbf{How mention is measured, in both directions.} Whether an explanation names the cue is
decided by a detector of five word bounded patterns over the generated text, and the reported
rates are adjudicated for precision and for recall rather than taken from the detector raw.
Every detector positive was read by hand, which removes false positives: on the primary model 35
of 41 hits were genuine, the six rejects all being one subtask whose own content concerns a
specialist. Every detector negative was then screened for false negatives through three
successive nets, a 44 term cue referring vocabulary far wider than the five detector patterns,
a net for attributions phrased without any of those five nouns, and an exhaustive person or
speech act screen of the residue matching neither, with every candidate read in each. Across all
8,464 explanations that search found 8 genuine mentions the detector had missed, and
7 of those 8 were censored by the 48 token generation cap rather than by vocabulary: the
explanation stops mid phrase immediately before the cue's noun while asking why the answer is
what it is. Truncation, not paraphrase, is the detector's failure mode here. The corrected rates
appear in Section 6, and recall is reported as measured rather than assumed because a
precision corrected count alone would be a lower bound on mention prevalence where the premise
needs an upper one.

\textbf{Models and compute.} The primary model is Qwen3-1.7B, with Qwen3-8B and Llama-3.2-3B
forming a replication tier reported per model and never pooled. All mechanistic work uses
instrumented forward passes with weights held in bfloat16 and every readout involving a difference
of logits or log probabilities computed in float32, since bfloat16's mantissa quantizes
exactly the small differences these readouts are designed to detect. Items are partitioned
once into nested development, selection and test splits by a hash of the item identifier,
with design choices fixed on development and selection and the test split analysis read
once on test.

%% file: sections/04_three_tests.tex
\section{Three estimators, run in the order the framework motivates}

Section 2 argued that three kinds of mechanistic evidence are routinely offered for a
causal access claim, that each rests on an inference from a signature compatible with use to
use itself, and that each has a distinct nuisance alternative a generic version of the same
statistic would also produce. This section says what was actually run. Appendix~E gives each
estimator in full, together with the development stage evidence that motivated it.

\textbf{Component attribution.} Ablate or patch components of the explanation pass one at a
time, rank them, and ask whether a sparse set carries most of the cue's effect on the
explanation readout. The favorable statistic is concentration, measured as the effect
recovered by the top $k$ components over that recovered by a matched random set of the same
size. The falsifying control is that matched random set, drawn to share layer and component
kind, plus a power gate requiring both the effect over spread and the top over random ratios
to clear a fixed bar on an independent subset. Without the power gate a concentration ratio
is uninterpretable, because a ratio near one can mean either that the circuit is diffuse or
that the measurement lacks the resolution to see it.

\textbf{Whole component transport.} Patch the decision pass's own component states into the
explanation pass and ask whether the explanation readout moves. The favorable statistic is
that movement. The nuisance alternative is that any foreign activation of the right shape
moves it, so the falsifying control is a donor identity contrast: a donor carrying the
counterfactual cue is patched in alongside one carrying the true cue, and the effect is
decomposed into a generic component, present for both donors, and a cue specific component
that depends on which cue the donor carried. Two further controls guard the construction. A
token length matched neutral prefix replaces the cue rather than deleting it, since deleting
it changes sequence length and misaligns positions under rotary embeddings, which was the
defect that invalidated an earlier version of this estimator. An invalid option sham donor,
naming an option that does not exist, shows what a purely generic effect looks like.

\textbf{Recoverable cue representation.} Fit a direction that predicts cue identity from the
explanation pass out of fold, then intervene along it. The favorable statistic is
out of fold reconstruction quality, and the nuisance alternatives are that the direction is
decodable but inert, and that the site was chosen for literal encoding rather than for use.
The falsifying controls are two null directions, one fitted by the identical pipeline with
cue labels scrambled and one drawn to match the activation covariance. This is the
primary estimator, and Section 5 specifies it and its decision rule in full.

An earlier single axis version of this estimator is preserved in the protocol as a failure. It does not survive its fold boundary, because a multiclass
variable has no single axis, and that failure is what motivated the multiclass construction.

The three subsections of Appendix~E share one template, apparent positive evidence, the
falsification control, what the control found in development, and the conclusion the evidence
permits. The three failures are instances of one gap, exposed by three different controls
because the three estimators face three different nuisance alternatives.

%% file: sections/05_methods_e5b.tex
\section{The primary estimator and its decision rule}

Every design choice in this section was frozen in the protocol before any
test data was seen. Appendix~B gives the full specification, including the
exact null construction, the cross fitting scheme, the numerical floor and the sample size
arithmetic.

\subsection{The model of cue identity}

The cue is a multiclass variable, since an item's cued option and its counterfactual option
are drawn from up to seven labels. A single binary axis cannot encode it, which a failed
earlier estimator established directly. We therefore model the state at a chosen site as
an item effect plus an option effect,
\[
h_{ic} \;=\; \alpha_i + \beta_c + \varepsilon_{ic},
\]
so that the paired contrast for item $i$ differs out the item effect,
\[
d_i \;=\; h_i(X_i') - h_i(X_i) \;=\; \beta_{X_i'} - \beta_{X_i} + \varepsilon_i.
\]
Stacking the items gives an incidence matrix $A$ whose rows carry $+1$ and $-1$ in the two
option columns, and the option effects follow as $\hat B = A^{+} D$ by pseudoinverse, which
is zero sum by construction. Option pairs are estimable only inside a connected component of
the incidence graph, checked by union find, and items whose contrast is not estimable are
reported and dropped rather than regularized in. The direction applied to a held out item is
pair conditional,
\[
v_{X \to X'} \;=\; \operatorname{norm}\!\big(\hat\beta_{X'} - \hat\beta_{X}\big).
\]

\subsection{Site, interventions and endpoint}

The intervention site is the residual stream at the last token of the decision prompt, the
position at which the answer is emitted. Within the explanation sequence that token is
followed by the answer line and the question asking why, so an edit there modifies a
decision side state that the later explanation computation reads through attention. The
mechanistic endpoint is a different position: it is read at the final token of the
explanation prompt, immediately before the explanation is generated. Keeping the two apart
matters, since the estimator asks whether editing what the decision left behind changes what
the explanation then does, not whether an edit made immediately beforehand does.

The layer is chosen by out of fold $R^2_{\text{cue}}$ on a development split, never by any
intervention outcome, and is then supplied unchanged to every later split. Selecting a site
by how well the cue is recoverable there finds where cue identity is most literally encoded,
and \S8 treats that as a property of the criterion rather than as a result.

Two interventions are run, both rank-1 and both at that single chosen position. Necessity
erases the component of the state along the direction, $h \leftarrow h - \langle h, v\rangle v$.
Interchange replaces it with the projection the state takes under the counterfactual cue.
There is no strength parameter to tune.

The mechanistic endpoint reads the next token distribution at the final explanation prompt
token under the clean and the intervened pass. With $u$ the natural cue swap contrast in log probabilities, the
signed readout is movement toward the counterfactual as a fraction of that swap,
\[
\Delta \;=\; \frac{\langle p^\ast - p,\; u\rangle}{\langle e^{\ell_{X'}} - p,\; u\rangle},
\]
so $\Delta = 1$ means the intervention moved the distribution as far as actually changing the
cue does. Necessity $N$ is $\Delta$ under erasure and interchange $S$ is $\Delta$ under
replacement. The observable endpoint is the greedily generated explanation, scored under both
contexts and held fixed, and \S2.5 fixes how its quantized cells are read.

\subsection{Nulls}

Two null directions are built for every held out item and carried through the identical
estimation and intervention procedure, so that any effect they produce is attributable to the
procedure rather than to cue identity.

The \textbf{shuffled label} null draws a fresh permutation of the option indices per training
item and applies it to both the cued and the counterfactual index, then refits the option
effects to the same contrasts. It passes through the same estimator, on the same items, at the same site, with the same
vector dimensionality, the same fitting procedure, the same normalization and the same
intervention operator. It does not preserve the option transition graph, and therefore not
the identification structure or the effective rank that follows from it, so we do not claim
equal degrees of freedom.

Because each item is relabeled independently, the shuffled training set spans
many more distinct option pairs than the target does and its degree profile is far from the
target's. We audited this directly, without touching any activation or endpoint, over two
hundred realizations per fold on each test split. The shuffled graph is uniformly
\emph{more} connected than the target's: it forms a single component in every realization,
whereas the target graph forms one to three, and in no realization on any split did the
shuffled graph fail to connect a held out item's own option pair. Identification is therefore
never weaker for the null than for the target, so the comparison is not one in which the null
is handicapped by unidentified contrasts. The difference in graph density is a structural
difference the construction does not control, and we report it as such.

The \textbf{covariance matched} null is a random direction drawn from the Gaussian with the
empirical covariance of the states at the site, then normalized, which matches the target in
dimensionality, spectrum and normalization.

The two answer different questions. The covariance matched null asks whether any direction of
typical activation geometry moves the readout, whereas the shuffled label null asks whether
the fitting pipeline alone, stripped of cue identity, produces the same effect. The second is
the stricter test, and \S6 shows it is the one that discriminates.

\subsection{Inference}

Items are assigned to five folds and every direction applied to a held out item is fitted on
the other four, so no item contributes to its own direction. The cue estimate is
deterministic given the items, and the seed enters only through the null realizations. The
protocol fixes three seeds as one test split analysis rather than three sequential
reads. Within each realization, inference is a percentile bootstrap over items with 10{,}000
resamples. Across realizations the nulls are summarized by mean and range. No aggregation
rule across seeds was specified and none is applied.

\subsection{The decision rule}

The verdict is not read off the point estimates. It is fixed by two gates and one access
criterion, all written into the protocol before the test split was opened
and all computed identically for every model, seed, intervention and endpoint.

\textbf{Gate A, variable specificity.} With $\theta$ the cue direction's effect and $\theta_0$
the null direction's effect on the same items, the specificity ratio is
\[
R \;=\; \frac{\lvert \theta_0 \rvert}{\lvert \theta \rvert},
\]
the share of the apparent effect that a direction carrying no cue identity already
reproduces. Gate A passes when $\mathrm{UCI}_{95}(R) < 0.50$, where the interval is a
percentile bootstrap over items with the cue and null effects recomputed together on each
resample so the pairing is kept. The threshold bounds the nuisance share rather than
significance: at $R = 0.50$ half the effect is delivered by a direction that knows nothing
about the cue, which is too much to call the evidence variable specific. Taking the upper
bound means a ratio that is small but poorly resolved does not pass.

\textbf{Gate B, practical zero.} A ratio is unstable when its denominator is near zero, which
is exactly the regime a null result produces, so Gate B asks the same question on an external
scale. It passes when $\mathrm{CI}_{95}(\theta_0) \subset [-0.10, +0.10]$ on the $\Delta$
scale above, where 1 is the movement produced by actually changing the cue. It is a check on
the comparison's validity rather than evidence for access, and it is reported whether it
passes or fails.

\textbf{Reading the gates in order.} The criterion already requires the cue direction's own
interval to exclude zero, and that requirement is read \emph{first}. When the target effect is
not resolved, the specificity ratio has a near zero denominator and is not stable enough to
carry an interpretation. It is still computed and reported exactly as specified, but it is not then treated as separate evidence of nonspecificity, since saying a model failed both
because its effect was unresolved and because its ratio was large would count one fact twice.
The ratio becomes substantively informative only once there is a resolved effect for a null to
reproduce.

\textbf{The access criterion.} A model is recorded as satisfying the variable specificity
criterion only when three things hold in the same realization: the cue direction's own interval excludes
zero, Gate A passes, and Gate B passes. Any one failing is a failure of the criterion. Satisfying it is necessary for the causal
reading of \S2.4 and is not sufficient for it, which is why the magnitude audit below and the
identity and agreement limitation of Section 3 are reported beside every pass. It is
never satisfied by a combination across seeds, endpoints or interventions, since necessity
and interchange answer different questions. No threshold was chosen or moved after any
test split number was seen.

\subsection{Realized edit magnitude}

A unit direction fixes its own length, not how much of the state it removes or replaces,
since erasure takes out exactly the component already lying along the direction and that
projection differs between directions even at equal norm. We therefore audit the size of the
edit each intervention actually made, reporting $m_N = \lvert \langle h, v \rangle \rvert$ for
erasure and $m_S = \lvert \langle h_i(X') - h_i(X),\, v \rangle \rvert$ for replacement, with
paired null to cue ratios by model, seed and intervention. The audit is diagnostic. It defines
no threshold and it rescales, reruns and reclassifies nothing. Its purpose is to say whether
a given comparison can be attributed to cue identity or is instead confounded by an imbalance
in how much was edited.

%% file: sections/06_results.tex
\section{Results}

Every number carries its split and its $n$. Only the three seed test split is read once, and its numbers were entered only after
the full three seed analysis existed, in the fixed reading order, never from a single seed. Per model detail is in
Appendix~C.

\subsection{The cue drives the decision and is rarely verbalized}

Table~\ref{tab:behavioral-gate} reports the behavioral gate on the full eligible pool,
$n = 2{,}116$ items, one run per model.

\begin{table}[h]
\centering
\footnotesize
\caption{Behavioral gate, authority channel, full pool. Mention is the adjudicated
mention rate, minority and required are the counted and needed minority arm
sizes, and power is design power. Usable $n$ is smaller than the 2,116 eligible items because
it counts only items on which the cue moved the answer and the explanation did not name the
cue, which is the population the mechanistic tests are defined on.}
\label{tab:behavioral-gate}
\begin{tabular}{lcccccc}
\toprule
model & flip rate & mention & $n_{\text{usable}}$ & minority & required & power \\
\midrule
Qwen3-1.7B & 68.3pp & 1.75\% & 1811 & 201 & 154 & 0.969 \\
Llama-3.2-3B & 63.8pp & 0.43\% & 1680 & 336 & 296 & 0.940 \\
Qwen3-8B & 65.0pp & 6.43\% & 1499 & 389 & 429 & 0.860 \\
\bottomrule
\end{tabular}
\end{table}

The cue raises the rate of answering the wrong option it names by 64 to 68 percentage points
against a paired uncued baseline in all three models, while the explanations
attribute the answer to it in under 2\% of items for three of the four models tested and
6.4\% for the fourth. Design power is the simulated power of the mechanistic test at the counted
size of the cue resisted arm, and it is not the test itself. Qwen3-8B falls short of the
 0.90 target at 0.860 and is carried forward with that shortfall reported.

Conditioned on whether an item followed the cue, the adjudicated mention rate is 1.84\%
against 1.12\% on Qwen3-1.7B, 0.53\% against 0.00\% on Llama-3.2-3B, and 3.12\% against
16.70\% on Qwen3-8B. Articulated disagreement with the cue therefore appears strongly in one
of the three models. However, the ordering by size is nonmonotonic, and with two
Qwen models and one Llama model neither scale nor family is supported as the
determinant. This is exploratory
and no explanation is offered. Mention status is posttreatment and is conditioned on
nowhere in the mechanistic analyses.

\subsection{Cue related structure is localizable and recoverable}

In earlier development work on a different cue channel, the explicit channel rather than the
authority channel used everywhere else here, the same attribution procedure localized a strong
decision side effect: the top 24 of 476 components recovered 91\% of the cue's effect at
2.85$\times$ a layer and kind matched random baseline (Qwen3-1.7B, $n = 35$). Its role is a
positive control rather than a reference condition. It shows the machinery can localize a large
known decision side effect, and because it comes from a different manipulation it is not the
baseline against which the authority channel results are judged.

In the explanation relevant state, out of fold reconstruction of the paired contrasts by the
multiclass model reaches $R^2_{\text{cue}}$ of $+0.938$ on the test split for
Qwen3-1.7B at layer 3, $+0.972$ for Qwen3-8B and $+0.955$ for Llama-3.2-3B at layer 2. Cue
identity is therefore recoverable from the explanation pass to a high standard. The retired
single axis rule's layer scores $+0.004$ on development, which independently confirms that
its earlier selection was an artifact of one item dominating an unweighted mean. A single
global direction does not survive its fold boundary at all, since 20 development items carry
14 distinct option pairs, so a multiclass variable has no single axis. This is the first
fact about representation the paper establishes and it holds independently of everything
below.

\subsection{Three estimators, three falsifications}

\textbf{Component attribution.} On the explanation pass the single position
KL readout localizes at top over random 2.42 on one item subset, whereas the
power gate fails on both of its criteria, effect over spread at 0.57 and top over random at
1.85, each against a required 2.0. Rescoring the answer
pass's own construct over the full sequence collapses it from 2.85 to 1.28, with one random
component reproducing 104\% of the full effect. Concentration is reproducible where the
signal is large and dissolves into generic component importance where it is small.

\textbf{Whole component transport.} Patching
decision circuit states into the explanation pass moves the readout by 0.194 for a native
donor and 0.189 for a counterfactual one. That the two are equal is what matters. With a
token length matched neutral prefix the donor identity contrast is
$-0.0017$ $[-0.041, +0.022]$, indistinguishable from zero at an interval 37$\times$ its point
estimate, whereas the generic component is clearly nonzero. An invalid option sham control shows what
a purely generic effect looks like, manufacturing a donor effect ten times the real one with
no cue specificity at all. The interval still admits a true cue specific component of about
$\pm2\%$ of the natural cue effect.

\textbf{Recovered cue representation.} This is the primary estimator,
and Table~\ref{tab:cross-model} gives its result on the three models, with the
additional model shown separately beneath them.

\begin{table}[h]
\centering
\footnotesize
\caption{Cross model summary, test splits, gates computed identically throughout.
Llama-3.2-1B was run after the other three, and it is the one case in which the criterion is
satisfied.}
\label{tab:cross-model}
\setlength{\tabcolsep}{3.5pt}
\scriptsize
\begin{tabular}{l p{2.3cm} p{2.3cm} c p{1.2cm} p{1.2cm} p{2.4cm}}
\toprule
model & cue $N$ & cue $S$ & cue excess (of 12) & Gate A & Gate B & failure mode \\
\midrule
Qwen3-1.7B & +0.0000 [$-0.0029$, +0.0030] & $-0.0002$ [$-0.0007$, +0.0003] & 0/12 & fails 12/12 & passes 12/12 & no resolved target effect, ratio not substantively interpreted \\
Qwen3-8B & $-0.000009$ [$-0.000145$, +0.000130] & +0.000116 [+0.000008, +0.000224] & 4/12 & fails 10/12 & passes 12/12 & signal too small, magnitude imbalance \\
Llama-3.2-3B & +0.0028 [+0.0022, +0.0035] & $-0.0000$ [$-0.0003$, +0.0003] & 5/12 & fails 11/12 & passes 12/12 & signal not specific against shuffle \\
\midrule
\multicolumn{7}{l}{\emph{Additional model:}} \\
Llama-3.2-1B & $-0.0006$ [$-0.0013$, +0.0001] & +0.00047 [+0.00020, +0.00075] & 6/12 & \textbf{passes 6/12} & passes 12/12 & criterion satisfied on interchange, audited magnitude confound \\
\bottomrule
\end{tabular}
\end{table}

The four models exhibit distinct failure modes, and the same fixed standard resolves each
differently. We therefore set them out separately.

\textbf{Qwen3-1.7B: no resolved target effect.} The cue direction's own effect is $+0.0000$
with an interval spanning zero. By the reading order of \S5.5 there is nothing for a null to
reproduce, so the specificity ratio is reported as specified and not interpreted
substantively. Where a null's interval does exclude zero, its sign is opposite the direction
use would predict. The operative fact is the absence of an effect rather than a failure of
specificity.

\textbf{Qwen3-8B: an effect resolved but negligible, and confounded.} The interchange effect is
statistically resolved, but its magnitude is about one hundredth of one percent of the
natural cue swap, and Gate A fails with the ratio on necessity running 12.6 to 53.8. The
realized edit magnitudes also favor the cue on the necessity nulls by 1.5 to 3.5 times. This
case shows that statistical resolution alone settles nothing, and it is why Gate B exists on an
external scale.

\textbf{Llama-3.2-3B: a substantial effect that is not specific, and the genuine
falsification.} It carries the largest cue effect measured
anywhere in this work, a necessity effect of $+0.0028$ $[+0.0022, +0.0035]$, about 0.28\% of
the natural cue swap, from a direction recoverable at $R^2$ 0.95. It beats the
covariance matched control with an interval excluding zero in all three seeds. However, the
shuffled label null reproduces 61 to 76\% of that same effect, around 70\% as a descriptive
summary, against a required upper confidence bound below 0.50. The two directions come from the same estimator at the same site under the same
normalization, and the audit in Section 5 shows the realized edit magnitudes are comparable,
with a median null to cue erase ratio of 0.83 to 1.04. However, independent scrambling also
changes the option transition graph, so this comparison is a strong falsification stress
test rather than an isolation of cue identity alone. A geometry only control would not have
rejected the result, whereas the label scrambled one does, which is why we report it as the
control worth reusing.

\subsection{What the realized edit sizes show}

Unit normalizing a direction fixes its length, not the size of the edit it induces, so every
comparison above is checked against the edit each intervention actually made.

The audit supports the comparability of the most informative comparison and weakens two
others. On Llama-3.2-3B the
shuffled label comparison is magnitude comparable, with a median null to cue erase ratio of
0.83 to 1.04, so the specificity failure there is not explained by a larger null edit. By
contrast the covariance matched null on that model received about 1.9$\times$ the cue's erase
magnitude, and on Qwen3-8B the two nulls received 1.5 to 1.7$\times$ and 3.3 to 3.5$\times$,
so those comparisons are confounded in a direction that can inflate the null effect. The imbalance
reverses on interchange for every model, where both nulls receive well under half the cue's
realized edit, which means favorable looking cue interchange comparisons cannot cleanly
establish specificity either. The audit is read as a validity check in both directions, not
as protection for our own reading.

\subsection{The observable endpoint}

At the rank-1 strength the intervention rarely crossed a greedy decoding argmax
boundary, so most free form cells are exactly zero. On the test splits the exact zero counts run 1126 of 1164 for Qwen3-1.7B, 2975 of 2988 for Qwen3-8B and 1987 of 2010 for
Llama-3.2-3B in seed 0, with the other seeds within a few cells. Where text does change under
a null it changes identically under both nulls, so the changes that exist are not
direction specific. This is the resolution property anticipated in \S2.5 and not an
equivalence result. An interval of $[0, 0]$ here is quantization, the gates are computed on
the mechanistic endpoint only, and the intervention strength was not raised after liveness was
confirmed once by a deliberately large edit.

\subsection{Summary}

\begin{table}[h]
\centering
\footnotesize
\caption{Each estimator's favorable statistic and the control that falsified the inference
built on it.}
\label{tab:estimator-summary}
\setlength{\tabcolsep}{4pt}
\begin{tabular}{p{2.5cm} p{3.1cm} p{2.9cm} p{3.1cm} p{2.0cm}}
\toprule
estimator & favorable evidence & specificity control & outcome & conclusion \\
\midrule
component attribution & top-$k$ concentrates the effect (2.85$\times$ on the decision pass) & matched random components, power gate & fails at 0.57 and 1.85 against 2.0 & concentration $\not\Rightarrow$ access \\
whole component transport & donor changes the readout (0.19) & donor identity contrast, neutral against invalid sham & $G \neq 0$, $C \approx 0$, $\Delta_{\text{ID}} = -0.0017$ $[-0.041, +0.022]$ & generic perturbation $\not\Rightarrow$ transport \\
multiclass cue direction & $R^2_{\text{cue}}$ 0.91 to 0.97 out of fold & shuffled label and covariance matched nulls & Gate A fails on all three models, passes on interchange in the fourth model & recoverability $\not\Rightarrow$ use in the main model set \\
\bottomrule
\end{tabular}
\end{table}

Across the three models the variable specificity criterion is not
satisfied, and each fails it for a different reason. We do not attribute the differences to scale,
architecture or family. Scale does not account for the pattern, since the ordering across
1.7B, 3B and 8B is nonmonotonic with the largest model showing the smallest effect. Family
is not claimed either, since the set holds two Qwen models and one Llama model and family is
confounded with tokenizer, training data, instruct tuning and architecture. Descriptively,
and with no cause attributed, the two Qwen models behave alike whereas Llama differs. Results
are reported per model and never pooled.

The replication set is complete. All three models have
test split analyses, and no further estimator is added.

\subsection{A fourth model, and the one result that satisfies the criterion}

A fourth model, Llama-3.2-1B, was run after the other three. Its site rule, nulls, endpoints,
gates, thresholds, seeds and sample size are the same as theirs, with the layer chosen on its own
development split rather than inherited, and its test split was analysed once like the others.
Appendix~C.5 gives it in full. It contributes two things that support the argument and one that
complicates it.

\textbf{Recoverability does not track behavioral causal strength.} The cue moves the 1B's
answer on 29.7 percent of the full pool against the 3B's 63.8, while out of fold
recoverability is essentially unchanged, $R^2_{\text{cue}}$ of 0.941 against 0.955, at the same
layer chosen independently by the same rule. The two quantities are measured on different
samples, the behavioral rate on the full pool and the reconstruction on the test split,
so this is a comparison of two models rather than of two subsets. Cue identity stays exactly
as decodable while its grip on behavior weakens by half. We read this as a size associated
difference within one model generation, not as a demonstrated effect of scale, since two
checkpoints differ in more than their size and two points are not a trend.

\textbf{A favorable selection estimate did not persist.} On the selection split the 1B's cue
necessity effect was $+0.0018$ $[+0.0003, +0.0035]$, an interval excluding zero with neither
null resolved. On the test split the same quantity is $-0.0006$ $[-0.0013, +0.0001]$ and
spans zero. This is the concrete form of the argument in \S2 for fixing the reading procedure
in advance.

\textbf{The criterion is satisfied once, on interchange.} The 1B's cue interchange
effect is $+0.00047$ $[+0.00020, +0.00075]$, its excess over both nulls excludes zero in all
three seeds, Gate A passes in all six interchange realizations and Gate B passes in all
twelve. By the criterion fixed in \S5.5 this is a pass, and it is reported as one.

\textbf{A separate audit weakens its causal reading.} The realized edit magnitudes are not
comparable on interchange. Across the three seeds the paired null to cue median runs 0.355 to 0.385 for the shuffled
label null and 0.077 to 0.087 for the covariance matched null, so the cue direction receives
roughly three and twelve times the edit its controls receive. On erasure the same medians run
0.88 to 1.23 and 0.83 to 0.97, so the necessity comparison is magnitude comparable and clean. A cue intervention that is substantially larger than its
controls is an ordinary alternative explanation for a cue effect that exceeds them, and this
comparison therefore does not isolate cue identity from intervention magnitude. The audit
defines no threshold and reclassifies nothing: the criterion passes, but the magnitude
imbalance prevents a clean causal interpretation.

No post result rescaling or reclassification was performed. A design that equates the two amplitudes by construction is possible and belongs to a separate study.

%% file: sections/09_related_work.tex
\section{Related Work}

Four literatures share vocabulary with this paper but ask different questions or license
different inferences, and a fifth body of work sits close enough that we state the overlap
outright.

\textbf{Explanation and chain of thought faithfulness.} \citet{turpin2023unfaithful} show
that chain of thought explanations can be systematically silent about a biasing feature that
demonstrably drove the answer, and we borrow exactly that construction as our controlled cause.
\citet{lanham2023measuring} intervene on the chain of thought itself, which is a causal
question about the decision pass conditioned on generated text rather than about the
explanation pass conditioned on a fixed answer. \citet{chen2025reasoning} extend the program
to reasoning models and report that hint usage is verbalized in a minority of cases, again
read off the transcript. \citet{jacovi2020towards} supply the definitional scaffolding.
Two recent papers push this line toward causal language. \citet{matton2025walk} define
causal concept faithfulness and measure whether the concepts an explanation invokes are the
concepts that actually drove the answer, and \citet{chuang2026faithlm} evaluate explanation
faithfulness by intervening on the explanation and observing the effect on the prediction.
Both move past correlational scoring, though both still read the result at the level of
concepts and outputs rather than asking what the explanation's own forward computation does.
Every one of these measures faithfulness at the level of text and behavior: does the output
change when the purported cause is edited, does the explanation mention what it should. None
asks whether the explanation's own forward computation is causally influenced by the identity
of the cause, independent of what the text says. A model could pass every behavioral test
here while its explanation never reaccesses the cause, and conversely. Our cue is
deliberately not the object of study, as it is in that work. It is an experimentally controlled cause
whose effect on the decision is measured independently.

\textbf{Causal interventions in mechanistic interpretability.} Causal tracing and activation
patching \citep{vig2020investigating,meng2022locating}, attribution patching
\citep{nanda2023attribution,syed2023attribution}, circuit discovery
\citep{wang2023interpretability} and its automation \citep{conmy2023towards}, and
distributed alignment search \citep{geiger2021causal,geiger2024finding} supply every
intervention we run. A parallel critical literature shows these tools are more fragile than
their headline numbers suggest, since faithfulness scores are highly sensitive to the
corruption distribution and metric choice \citep{zhang2023towards} and reflect methodological
choices as much as the circuit itself \citep{miller2024transformer}. We extend this line by pairing each inference with a variable specific null that uses the same site and intervention procedure, with matched normalization or representational geometry where
the construction allows it, together with an audit of the edit magnitudes actually realized. Those critiques argue that metric and corruption choices
change the number, whereas we argue that even a well chosen metric can produce a favorable
number for a reason unrelated to the variable's identity, which is why the same nuisance
alternative recurs across all three of our estimators despite disjoint machinery.

\textbf{Introspection and self knowledge.} \citet{kadavath2022language} show models are
reasonably calibrated about whether they know an answer. \citet{binder2024looking} find a
model predicts its own counterfactual behavior better than an equally capable model does.
\citet{lindsey2025emergent} inject concept representations and ask whether self report
notices. All three ask whether a model's \emph{report} is accurate. Reporting accuracy is outside the present question, since the mention rate in our transcripts is a property of the setting and not a result. We ask whether an externally known and independently manipulated cause enters the
explanation's forward computation, whether or not it is ever named. Calibration, predictive
self modeling and detecting an injected concept are each compatible with either outcome of
that question.

\textbf{Decodability is not causality.} \citet{hewitt2019designing} show a probe's accuracy
must be compared against a control task of matched complexity.
\citet{ravichander2021probing} show high accuracy can arise from incidental presence rather
than use, and \citet{belinkov2022probing} surveys the resulting consensus.
\citet{elazar2021amnesic} move from decoding to intervention, \citet{belrose2023leace} give a
closed form for that removal, and \citet{geiger2021causal} arrive independently at causal
abstraction. Amnesic probing already compares learned removal directions with random
directions matched in number and tests selectivity by restoring the removed property.
Our representation test uses a shuffled label direction fitted by the same pipeline,
alongside a covariance matched null of the same norm and an audit of realized edit magnitudes.
We also apply this control logic to whole component
transport, where the nuisance alternative is not an inert direction but a donor whose effect
is independent of which cue it carries.

\textbf{The nearest work, and what remains different.}
\citet{makelov2024subspace} show subspace activation patching can produce an interpretability
illusion, where an intervention moves the output through a dormant parallel pathway rather
than the subspace it is attributed to. That is the same class of concern our transport
control addresses, arrived at independently. We test whether a transport effect depends on
which variable the donor carries, whereas they show the attributed subspace may not be the
route at all. \citet{yeo2025towards} apply activation patching to the faithfulness of natural
language explanations, which is the closest work on method, and it means no claim that we are
first to look inside activations when judging explanations would be correct. The differences
are in the experimental target and control structure. They construct counterfactual inputs
through symmetric token replacement and compare activation based causal attributions for
answers and post hoc explanations, including token, layer and multi-layer variants.
We measure an added cue's influence on the decision independently and test three
estimator classes, each against a null that removes cue identity while matching the
relevant nuisance structure as far as the construction permits. \citet{singh2026introspect} formalize
necessary conditions for claims of introspection and show earlier positive results fail under
input only baselines and relabeled controls. We regard it as a precedent rather than a
contrast. It targets self report of an internal state, where we target post hoc explanation of a
decision whose cause was injected and measured beforehand. \citet{sudheendra2026decodability} report the same shape of result for logical validity:
validity is almost perfectly decodable from hidden states even where behavior is near
chance, whereas interventions along the probe derived direction have weak and nonspecific
effects relative to random controls. They reach a related conclusion about logical validity, and their controls are random
directions rather than a null fitted by the same pipeline with the variable's identity removed.
Their subject is also a property the model may or may not compute, where ours is a cause we
inject and measure before the explanation is generated. \citet{tiwari2026decodability} dissociate what a probe can read from
what drives behavior using sparse autoencoder decomposition on a truthfulness probe. They audit safety probes where we audit self
explanation, and their evidence rests on one form of evidence where ours rests on three, each
with its own control.

\textbf{What is new here.} Prior work has separately shown that explanations can be unfaithful,
that interventions can test causal dependence, and that decodable information need not be
causally used. We claim none of those findings. What this paper adds is a variable specific null
principle applied systematically across three different kinds of mechanistic evidence,
attribution, cross pass transport and representation recovery, inside one system with an
injected and independently measured cause, with every control fixed before the test data
was read, and with an audit of what each intervention actually did to the state.
Three of the controls built for that purpose are reusable beyond this setting: a decomposition
of a cross pass transport effect into a generic component and a component specific to the transported variable's identity, reported separately, a null
direction fitted by the same pipeline with the variable's identity scrambled, standing beside
any geometry matched control rather than in place of it, and an audit of the edit sizes an
intervention actually applies, since normalizing a direction does not make the edits it
induces comparable.

%% file: sections/07_discussion.tex
\section{Discussion}

Each estimator produced a statistic that looks, on its own, like evidence for causal access:
a sparse set of components recovering most of an effect, a donor patch that visibly moves an
explanation, a direction whose out of fold $R^2$ approaches one. Each also failed the control
built for its own nuisance alternative. The pattern is not that three methods happened to
fail. It is that attribution, transport and representation recovery are progressively
stronger \emph{kinds} of evidence, whereas progressively stronger evidence is still not
sufficient, because each kind admits a cue agnostic explanation for the same number.

\textbf{Decodability is not use.} A reconstruction criterion asks where a variable is most
\emph{literally} encoded, and for a variable injected as tokens in the prompt that is early
and close to those tokens, since that is where the representation is least transformed.
Nothing in an $R^2$ criterion distinguishes a site that encodes the cue from one the
explanation reads. Layer 3 scoring highest is therefore a property of what reconstruction
criteria find, not evidence about where the cue is used. This generalizes past our estimator:
probing and feature finding methods are criteria for encoding rather than for causal
relevance, and the fix is not a better probe or a later layer, but refusing to let the
representational criterion double as the causal one.

\textbf{Transport controls.} A foreign activation can move a downstream readout for
reasons unrelated to its semantic content. Our
first transport run reported an effect of the expected size, later traced to positional
misalignment rather than to semantic transport, which rotary embeddings make fatal rather
than cosmetic. Once corrected, the donor identity contrast collapsed to indistinguishable
from zero, whereas an invalid option sham control manufactured a generic effect an order of magnitude
larger. The generic term can dwarf the cue specific one, so reporting a single patch effect
size, which is what ``the explanation changed when we patched in the decision pass
activations'' amounts to, says nothing about semantic content. Report the two separately and
license the claim only on the second.

\textbf{Specificity relative to nulls.} A confidence interval excluding zero does not by
itself establish specificity, since correlated nuisance structure can also produce a resolved
intervention effect. The relevant quantity is always the excess of the cue statistic
over a matched null computed through the identical estimation and intervention procedure, so
that shared artifacts cancel. Qwen3-8B makes this concrete: its interchange effect is statistically resolved and
still fails the specificity gate by one to two orders of magnitude.

\textbf{Realized intervention magnitude.} These
are the two transferable recommendations. On the model with the largest cue effect, a
direction fitted by the identical pipeline with cue identity scrambled reproduced most of
that effect, whereas the covariance matched null standing beside it missed this almost
entirely. Anyone running such an intervention should include a scrambled identity direction
and not only a geometry matched one. Separately, unit normalizing a direction fixes its
length as a vector, not the size of the edit it makes, since erasure removes exactly the
state's own projection and that projection differs between directions at equal norm. Read
against our audit, two necessity comparisons are magnitude comparable and two are not, in
both directions. A control described as the same site, the same operator and the same
normalization is not thereby matched on edit size.

\textbf{Heterogeneity without changing the criterion.} The same test, applied unmodified to
three models, returns three different outcomes: no detectable signal, a resolved but
negligible one, and a larger reproducible one that beats a geometry matched null but not the
stricter shuffled label null. The criterion was fixed before any test split analysis and
adjusted after none of them. The same test tells these cases apart rather than
returning one verdict regardless of what is in front of it. None of the three passes, so none
supports a causal access claim, whereas the way each fails is informative and is reported per model. A fourth model, added afterwards under the same fixed rules, does pass on
one endpoint, and we report that pass beside the magnitude audit that complicates it.

\textbf{What the pattern supports.} The ladder from Section 1 is the sum of these results.
The cue causally changes the decision, cue related structure is localizable and recoverable,
no estimator in the main three model analysis yields evidence of causal sensitivity to
the cue contrast surviving its control,
and therefore attribution, transportability and recoverability are not, individually or
together, sufficient evidence of causal use. In short,
\[
\text{represented} \not\Rightarrow \text{used}, \quad
\text{perturbable} \not\Rightarrow \text{semantically transported}, \quad
\text{significant} \not\Rightarrow \text{variable-specific}.
\]
The last step is asymmetric by construction. A passing result is necessary for a causal access
claim and does not on its own establish one, while each failing result supports only ``not established'' for the pathway that
estimator tested, never that no causal access exists, since three estimators however
differently constructed do not exhaust the space of pathways.

The free form endpoint sits alongside these results rather than adjudicating them. An
interval of $[0, 0]$ there is a resolution floor, not an equivalence statement, and it is
never used to rescue an estimator that failed its control or to be rescued by one that
passed.

We report one observation without proposing an explanation. Conditional on resisting the cue,
one of the three models articulates that disagreement at a rate far above the other two,
tracking neither scale nor family. We leave it as a checkable target and do not fold it
into the causal access argument.

These lessons are supported by a controlled setting where the cue is injected and its effect
on the decision independently established, so estimators can be checked against a positively
validated decision side reference. They are not claimed as universal properties of language
model explanations, and we do not extend them past the estimators and models tested here.

%% file: sections/08_limitations.tex
\section{Limitations}

Appendix~D states these in full, along with four further ones. Five matter enough to belong
here.

\textbf{The cue is artificial.} It is an injected authority statement naming a specific wrong
option, chosen so that its causal role in the decision could be established by construction
and by measurement. That buys certainty about what the cause is and that it drives the
answer, while a bias the model acquired during training may be weaker, spread across more
components, or not sit near a token span at all. Nothing here licenses extrapolating to that
harder case.

\textbf{The intervention family is narrow.} Every intervention is either a rank-1 edit at one
residual stream position or a whole component patch between matched passes.
Neither covers multi position interventions, nonlinear edits, or interventions on attention
patterns. A pathway needing coordinated changes across positions, or one not linearly
represented at the tested site, is invisible to every estimator here by construction,
independent of whether it exists.

\textbf{The site is chosen by recoverability.} The layer is the one where the cue is most
literally decodable out of fold, which \S8 argues will tend to be early and token proximal
regardless of where the cue is used, because the criterion cannot distinguish encoded from
used. A causally motivated rule would face the opposite problem the protocol was
built to prevent, since selecting a site by an intervention's outcome biases the specificity
test that follows. No such rule was specified, and none is retrofitted now that the result
is known.

\textbf{Explaining a forced answer is not explaining a chosen one.} The answer being explained
is held fixed at the cued option in every condition, so that an intervention can change only
the explanation. On items where the model followed the cue the prompt asks about an answer it
did give, while on items where it resisted the prompt asks it to justify one it did not
choose. The analyses pool both, since conditioning on which arm an item fell into would
condition on a posttreatment variable. Any reading of these results as being about
self report of a real choice applies cleanly only to the followed arm.

\textbf{The counterfactual moves two things at once.} Switching the cue from the named option
to the counterfactual one changes which option the cue names, but it also changes whether the
cue agrees with the answer being explained, since that answer stays fixed. The contrast
therefore carries cue identity and cue answer agreement together, and this design cannot
separate them.

\textbf{The conclusion is asymmetric, by the rule fixed in \S2.4.} Failure of these three
estimators to produce evidence of sensitivity to the cue contrast surviving their
controls does not establish that
the explanation computation lacks causal access to the cue. It establishes that these
interventions, at these sites, on these endpoints, at these strengths, did not produce it.
Pathways left open include a subspace other than the one tested, a nonlinear encoding no
linear or component level intervention would disturb, an explanation specific recoding of the
cue, and components outside the tested attribution and transport sets.

%% file: appendix/A_framework.tex
\section{Framework details}

This appendix gives the full operationalization behind Section 2. The main text states the
inequality, the specificity requirement and the reading rule. What follows is the longer
argument for each.

\subsection{What counts as causal access}

Fix a multiple choice item and a model. Three objects are in play.

\begin{itemize}
\item \textbf{$C$, the cause.} A feature injected into the prompt that is known, by construction and
  by measurement, to drive the answer. In this paper $C$ is a cue naming a deliberately
  wrong option. Section 3 establishes that it moves the answer to that option in a large
  majority of items. $C$ takes values in the option set, so it is a \emph{multiclass} variable:
  the question is never ``is there a cue'' but ``which option does the cue name''.
\item \textbf{$D$, the decision computation.} The forward pass that produces the answer $y$ under
  $C$. Because $C$ is known to drive $y$, $D$ demonstrably contains a pathway from $C$ to
  $y$, and that pathway can be localized (Section 4). $D$ is the measured reference the rest of
  the framework is measured against.
\item \textbf{$E$, the explanation computation.} The forward pass that, given the answered item,
  produces the free form text when the model is asked why it chose $y$. The paper's question
  is about $E$.
\end{itemize}

The claim under test is that \textbf{$E$ has causal access to $C$}: that the explanation
computation is causally influenced by \emph{the identity of the cue}, not merely by the answer
the cue produced or by the tokens the cue occupies. Three progressively stronger properties
have to be kept apart, because ordinary usage of ``the model knows why it answered'' slides
between them:

\[
\begin{aligned}
&\underbrace{C \text{ is decodable from } E\text{'s activations}}_{\text{present}} \\[8pt]
{}\neq{}\quad &\underbrace{\text{perturbing those activations changes } E\text{'s output}}_{\text{used}} \\[8pt]
{}\neq{}\quad &\underbrace{\text{changing } C\text{'s identity there moves } E\text{'s output toward } C'}_{\text{used specifically}}.
\end{aligned}
\]

Only the third is causal access. The first is a statement about the representational
content of $E$'s residual stream, which can be high for reasons that have nothing to do
with $E$'s output (the cue's tokens are in the context and are trivially readable early in
the network). The second is a statement about the causal relevance of a \emph{site}, which is
satisfied by any site that matters for fluent generation whether or not it carries cue
identity. The third requires that the effect on $E$ be a function of \emph{which} cue is present,
i.e. that substituting $C \to C'$ inside $E$ moves the explanation toward what it would have
been under $C'$, and that substituting an uninformative perturbation built under the same
construction does not.

We therefore define:

\begin{quote}
\textbf{Causal access.} $E$ has causal access to $C$ at a site $S$ if an intervention at $S$
that replaces the representation of $C$ with that of a counterfactual $C'$ changes $E$'s
output in the $C \to C'$ direction by more than a matched intervention that carries no
cue identity.
\end{quote}

Every term in that definition is operationalized in Section 5: the site, the replacement,
the direction, the endpoint on which ``changes'' is read, and the matched intervention. The
last clause is the one that separates this definition from the usual ones, and the rest of
this section is about why it has to be there.

\subsection{Why ordinary mechanistic evidence is insufficient}

Three kinds of evidence are routinely offered for causal access claims. Each is a real
measurement of something. None of them measures the third property above, because each has
a \emph{nuisance alternative} that produces the same number without cue specific use.

\textbf{Attribution concentration.} One ablates or patches components of $E$ one at a time,
scores the change in some readout, and reports that a small set of components carries most
of the effect. The inference is \emph{attribution $\to$ access}: the components that matter for
the readout are the components that carry $C$. The nuisance alternative is that the readout
is sensitive to \emph{generic} component importance. Every component that matters for producing
fluent text at all will register on any readout that depends on that text. On a weak signal,
the generic term dominates the cue specific one, and a sparse, high scoring set of components
is exactly what one expects to find whether or not any of them carries cue identity. The
reference distribution against which ``concentrated'' is judged has to be built from
components matched on their generic importance, not from components drawn at random. In
plain words: a sparse set of components can score highly because those components matter for
fluent text in general, not because they carry the cue's identity. The matched falsifying
control (\S2.3) is a set of components chosen for the same layer, kind and generic
importance, and the claim survives only if the true components beat that matched set.

\textbf{Transport effects.} One takes the activations of a set of components under $C$ from the
decision pass, patches them into the explanation pass, and reports that the explanation
readout moves. The inference is \emph{transport effect $\to$ semantic transport}: the decision
pass's representation of $C$ was carried into $E$ and used there. The nuisance alternative
is that patching any foreign activation into $E$ moves the readout. A donor activation is a
perturbation of a particular magnitude and geometry at a particular set of positions.
Positional misalignment alone can produce large effects, and so can any donor that differs
from the target for reasons unrelated to cue identity. What the inference needs is that the
effect \emph{depend on which cue the donor carries}: a donor from $C$ and a donor from $C'$ must
move the readout differently, and the difference is the transport effect. The effect common
to both donors is generic and must be subtracted, not reported. Put plainly, a donor
activation can move the readout because it is a foreign perturbation of a given size and
position, not because it carries the cue's identity. The matched falsifying control (\S2.3)
is a donor that carries a different cue, or a length matched neutral donor that carries no
cue at all, and the claim survives only if the true donor moves the readout more than these
do.

\textbf{Recoverability.} One fits a direction, a subspace or a probe that recovers $C$ from the
activations of $E$ at some site, reports its out of sample accuracy, and then intervenes
along it. The inference is \emph{recoverability $\to$ causal use}: the network represents $C$
here, so this is where $C$ acts. The nuisance alternative is that a recoverable
representation can be causally inert with respect to $E$'s output. Reconstruction criteria
select the site where $C$ is most \emph{literally} encoded, which for an in context cue is close
to its tokens and early in the network. Nothing in the criterion distinguishes a site that
encodes $C$ from a site that uses it. And the intervention along a recovered direction is a
perturbation like any other, so its effect must again be compared against a matched
perturbation that carries no cue identity. A direction that is decodable at $R^2$ near one
and that, when intervened on, moves $E$ by no more than a shuffled label direction of the
same norm has established presence and nothing further. A direction can be
reconstructed at high accuracy because the cue's identity is represented there, without that
representation ever being used by the computation that produces the explanation. The matched
falsifying control (\S2.3) is a direction fit to shuffled cue labels or matched to the
activation's covariance, and the claim survives only if intervening on the true direction
moves the explanation more than intervening on these does.

The three inferences share a structure. Each takes one of the commonly used empirical
signatures compatible with causal access, namely that a used variable tends to show up as
attributable, as transportable, and as recoverable, and treats that signature as
\emph{sufficient}. We do not claim any of the three is strictly required for use, since a
variable could be used through a route none of them would register. The point is the
reverse direction: each signature is routinely read as evidence of use, whereas the
measured effect behind it has a cue agnostic component, and only the cue specific component
bears on the claim.

%% file: appendix/B_methods.tex
\section{The primary estimator in full}

This section specifies the third and most targeted estimator of \S2.2, exactly as it was
frozen in the protocol (Sections 5b, 9b, 9c and 10) before any
test split was read. It is written to be applied unchanged to every model in the replication
tier: the only model specific quantity is the layer, which is selected by a
representational criterion on that model's development split and then held fixed. Nothing
in this section depends on an intervention outcome. The numbers it produces are reported in
\S6.

\subsection{Data and splits}

Items are drawn from the BIG-Bench Hard subtasks with inline multiple choice options, under
the authority cue channel (``A domain expert reviewing this question concluded that the
answer is (X).''), with the cue always naming a wrong option. An item is eligible if it has at
least two wrong options, so that a paired counterfactual cue $X'$ exists. The eligible pool
(2,116 items) is partitioned once into \textbf{development, selection and test} splits of 202 /
298 / 1,616 items, by the rank of a hash of the item identifier within each subtask, so that
every subtask is represented in every split in proportion and membership is a pure function
of the item identifier. The estimator, its layer, its nulls and its endpoints were developed
on the development split, run once on the selection split as a mechanical validity check,
and run once on the test split. These restrictions are enforced by the analysis pipeline rather than only by the written
protocol: the test split run requires an explicit invocation, will not repeat a seed
already completed, and closes after the three specified seeds (Appendix B.7).

For each item $i$ the design supplies a cued prompt with cue $X_i$, the same prompt with the
counterfactual cue $X_i'$, and the post hoc explanation prompt: the cued question, the
model's answer line naming $X_i$, and the line \emph{Why did you choose ($X_i$)? Because}. The
explanation pass is the forward pass over that prompt. The answer is held fixed at $X_i$ in
every condition, so an intervention can only change the explanation, never the thing being
explained.

\subsection{The model of cue identity}

The paired design identifies only \emph{differences} between option representations, and the
estimator is written in those terms. At the residual stream of one layer $L$ and one position
(Appendix B.3), the state under cue $c$ on item $i$ is modeled as an item effect plus an option
effect,

\[
h_{ic} = \alpha_i + \beta_c + \varepsilon_{ic},
\]

so that the paired contrast cancels the item effect exactly:

\[
d_i \;=\; h_i(X_i') - h_i(X_i) \;=\; \beta_{X_i'} - \beta_{X_i} + \varepsilon_i.
\]

With $A$ the $n \times K$ incidence matrix holding $+1$ at column $X_i'$ and $-1$ at column
$X_i$ in row $i$, and $D$ the $n \times d_{\text{model}}$ matrix of contrasts, the option
effects on a training set are the least squares solution

\[
\hat B \;=\; A^{+} D,
\]

where $A^{+}$ is the Moore Penrose pseudoinverse. Because $A^{+}$ returns the minimum norm
solution, which is orthogonal to the null space of $A$, and that null space on a connected
option graph is spanned by the all ones vector, the identification constraint
$\sum_c \beta_c = 0$ holds by construction. There is no rank choice, no regularization
parameter, no probe and no dimensionality reduction. The model has exactly as many free
parameters as there are options, and it pools information across items
(training contrasts $A \to C$ and $C \to E$ jointly inform the implied contrast $A \to E$)
without assuming that all contrasts lie on one axis.

This replaces a single global direction $v = \operatorname{norm}(\sum_i d_i)$, which was the
fixed construction in the protocol and which failed its own out of fold sign check on
the development split before any intervention was run. The reason is structural rather than statistical: the
cue names one of up to seven options, so cue identity is a multiclass variable, and a
binary axis cannot encode it. The single axis construction is preserved in the protocol as the failed
attempt, and the amendment was made at the one point at which it is defensible: with no
causal result in hand.

\textbf{Identifiability is reported, not regularized away.} A held out item requires the contrast
$\hat\beta_{X'} - \hat\beta_{X}$, which is estimable only if $X$ and $X'$ lie in the same
connected component of the option transition graph of the training fold. Components are
computed by union find over the training pairs. An item whose contrast is not estimable is
counted, reported, and dropped from that fold. It is never filled in by shrinkage, because a
direction the data cannot support is the same class of object as every other artifact this
paper is about.

\subsection{Site: layer and position}

\textbf{Position.} States are read, and interventions applied, at the residual stream at the
\emph{last token of the decision prompt}: the position at which the
answer is emitted. Inside the explanation prompt this token is followed by the answer line
and the \emph{why} line, so an edit there is an edit to the decision side state that the
explanation pass reads through attention. This is the sense in which the estimator is
decision defined: the direction is fitted on the decision pass and its causal test is run on
the explanation pass.

\textbf{Layer.} The layer is selected on the development split by out of fold reconstruction
of the paired contrasts,

\[
R^2_{\text{cue}}(L) \;=\; 1 - \frac{\sum_i \lVert d_i - (\hat\beta_{X_i'} - \hat\beta_{X_i}) \rVert^2}{\sum_i \lVert d_i \rVert^2},
\qquad \hat\beta \text{ fitted without item } i\text{'s fold},
\]

evaluated at every layer, and the best scoring layer is frozen before the selection or test
splits are touched.
The criterion is a property of the representation and never consults an intervention
effect, so a layer selected by it cannot have been selected because an intervention worked
there. The layer chosen for the discarded single axis estimator is \emph{not} inherited: once that
estimand was abandoned, a selection made under it had no standing. Reconstruction criteria
have a known bias that \S2.2 anticipates (they select the site where cue identity is most
literally encoded, not the site where it is most causally relevant), and the framework
treats that as a limitation to report, not a reason to search other layers after the fact.

\subsection{Interventions}

For each held out item the pair conditional direction is

\[
v_{X \to X'} \;=\; \operatorname{norm}\!\big(\hat\beta_{X'} - \hat\beta_{X}\big),
\]

fitted on the other four folds. Two interventions are applied at the site, both through the
same intervention operator, which receives a unit vector and is blind to which condition it
is serving:

\begin{itemize}
\item \textbf{Necessity (erase).} The component of the state along $v$ is set to zero:
  $h \leftarrow h - \langle h, v\rangle\, v$. If $E$ uses the cue identity component along
  $v$, removing it should move the explanation \emph{away} from the $X$-consistent reading.
\item \textbf{Interchange (replace).} The component along $v$ is set to the scalar projection the
  state takes on the \emph{natural} $X'$-cued run of the same item:
  $h \leftarrow h + \big(\langle h_i(X'), v\rangle - \langle h, v\rangle\big)\, v$. If $E$ uses
  the component, the explanation should move \emph{toward} the $X'$-consistent reading.
\end{itemize}

Neither intervention has a strength parameter: erase removes exactly the component present,
and replace installs exactly the value observed under the counterfactual. Nothing is scaled
up, and per \S2.5 nothing is scaled up later.

\subsection{Nulls}

Two null directions are built for every held out item and pushed through the identical
estimation and intervention procedure (incidence construction, pseudoinverse,
normalization, and both interventions) so that any effect they produce is attributable to
the procedure and not to cue identity. What each null does and does not match is stated below.

\begin{itemize}
\item \textbf{Shuffled identity.} The training fold's items are relabeled, the option effects
  are refitted to the \emph{same} contrasts $D$, and the held out item's direction is formed
  from the refitted effects. The null passes through the same estimation and intervention
  procedure, on the same items, at the same site, with the same
  vector dimensionality, the same fitting procedure, the same normalization and the same
  intervention operator, and differs in that its option labels carry no information about
  which cue was present. It does not preserve the option transition graph, so we claim
  neither an identical identification structure nor equal degrees of freedom.

  The construction is the following, stated exactly as it runs. For each \emph{training} item,
  draw a fresh permutation $\pi$ of the option indices $\{1, \dots, K\}$ from the seeded
  generator, then set that item's cued index to $\pi(X)$ and its counterfactual index to
  $\pi(X')$. The activations, the contrast $D$, the fold assignment and the item order are
  untouched. The refit then runs the same incidence construction, the same pseudoinverse and
  the same normalization as the target. The held out item is \emph{not} relabeled: its
  direction is read from the refitted effects at its own original pair $(X, X')$, so the null
  and the target are evaluated at exactly the same contrast and differ only in the option
  effects that contrast is formed from. Because the relabeling is bijective and
  $X \neq X'$, label collisions cannot occur.

  \textbf{What the relabeling does not preserve.} Because $\pi$ is drawn per item rather than
  once per fold, the shuffled training set does not inherit the target's option transition
  graph. We audited this directly, on the item sets alone, with no activation, endpoint or
  outcome involved, over 200 independent realizations per fold on each test
  split. The shuffled training set spans roughly three to four times as many distinct
  option pairs as the target (for example, a median of 181 against the target's 46 to 50 on
  the $n = 340$ split), and its degree profile sits at a total variation distance of about
  0.62 from the target's on every fold of every split. What it does preserve is the
  property that matters for identification, and it preserves it in the favorable direction:
  the shuffled graph formed a \emph{single} connected component in every one of the 3,000
  realizations audited, against one to three components for the target, and in no realization
  did it fail to connect a held out item's own option pair. The null's directions are
  therefore always identified contrasts, never minimum norm solutions of an underdetermined
  system, so the comparison is not one in which the null is handicapped. The difference in
  graph density is a structural difference the construction does not control for, and it is reported as one.
\item \textbf{Covariance matched.} A random direction drawn from the Gaussian with the empirical
  covariance of the states at the site, $\Sigma = \operatorname{cov}([h(X); h(X')])$, then
  normalized. It matches the target in dimensionality, spectrum and normalization, and is
  the control for the possibility that any direction of typical activation geometry moves
  the readout.
\end{itemize}

Both nulls are realized from a seeded generator (Appendix B.7). Per \S2.3 the specificity claim is the
excess of the cue direction over each null, read separately for necessity and interchange.
The two claims are never combined.

\subsection{Endpoints}

\textbf{Mechanistic endpoint.} At the last token of the explanation prompt (the position at
which the first word of the explanation is emitted), the next token distribution is read
under the clean pass, $p$, and under the intervened pass, $p^\ast$. Let $\ell_X$ and
$\ell_{X'}$ be the clean log probabilities at that position under the $X$-cued and
$X'$-cued explanation prompts, and $u = \ell_{X'} - \ell_X$ the natural cue swap contrast.
The signed readout is the movement toward $X'$ as a fraction of the natural swap,

\[
\Delta \;=\; \frac{\langle p^\ast - p,\; u\rangle}{\langle e^{\ell_{X'}} - p,\; u\rangle},
\]

so that $\Delta = 1$ means the intervention moved the distribution as far toward the
$X'$-cued reading as actually changing the cue does. It is the same readout used by the
transport estimator, so the two are on one scale. Necessity $N$ is $\Delta$ under erase.
Interchange $S$ is $\Delta$ under replace. Items whose denominator is numerically zero are
reported as nonevaluable, not imputed.

\textbf{Observable endpoint.} This is the endpoint the protocol designated primary,
and we keep that designation, because what happened to it
is part of the record. It turned out to be resolution limited at the intervention
strength, so almost every cell is exactly zero, whereas the gates and every inferential
conclusion in this paper are computed on the mechanistic endpoint. A reader is entitled to
ask how the nominated primary endpoint can carry no inference. The answer is that we
discovered its resolution floor after registering it and before reading any
test split result, and that the response to a quantized endpoint is to report it as quantized. Relabeling the
mechanistic endpoint as primary after the fact would have been the easier presentation, and we
have not done it.

The explanation is generated greedily for
48 tokens from the clean prompt and from the prompt under each intervention, the
intervention applied on the prompt pass and therefore cached into every subsequent generation step. Each
generated text $t$ is scored, held fixed, under the two contexts,

\[
z(t) \;=\; \log P(t \mid X'\text{-cued explanation prompt}) - \log P(t \mid X\text{-cued explanation prompt}),
\]

summed over the continuation and not length normalized (both terms score the same string,
so length is identical between them). The causal quantities are
$N_{\text{ff}} = z(t_{\text{clean}}) - z(t_{\text{erase}})$ and
$S_{\text{ff}} = z(t_{\text{replace}}) - z(t_{\text{clean}})$. Under greedy decoding an
intervention that crosses no argmax boundary yields $t_{\text{erase}} = t_{\text{clean}}$
and an exact zero. \S2.5 fixes how such cells are read. Intervention liveness was verified
once, before any run, by verifying that a deliberately large edit at the same
site changes the generated text. The interventions are not scaled toward that
magnitude.

\subsection{Cross fitting, seeds and inference}

Items are assigned to five folds by index modulo five. For each fold the option effects,
the connectivity structure and the shuffled identity effects are fitted on the other four
folds. Every direction applied to a held out item is therefore out of fold, and no item
contributes to its own direction.

The cue estimate is deterministic given the items. The seed enters only through the null
realizations (the label permutation and the covariance matched draw), and Section 10 of
the protocol fixes \textbf{three seeds, $\{0, 1, 2\}$}, as one test split analysis.
Within each realization, inference is a percentile bootstrap over items (10,000 resamples)
of the mean of $N$ and of $S$ for the cue direction and for each null. Across the three
realizations the null estimates are summarized descriptively by their mean and range. No
aggregation rule across seeds was specified and none is applied. The primary inference is
within each realization.

\textbf{Sample sizes.} The development and selection splits were run at $n = 40$ and $n = 60$.
The test split's sample size, fixed before that split was accessed, is $n = 200$: the
design power minimum of 154 usable items for the primary model, inflated by the
nonestimable rate observed on the development split ($154 / (1 - 0.125) = 176$) and rounded
up. If fewer than 154 contrasts prove estimable that is reported. The sample is not extended.

\textbf{Reading order.} The test tables are read in the order fixed in the protocol and stated in
\S2.5: (1) integrity: manifest, fold partition, cell counts, nonestimable and nonevaluable
counts, (2) $R^2_{\text{cue}}$ as representation validation only, (3) cue $N$ and $S$,
(4) shuffled and covariance matched $N$ and $S$ for all three seeds, (5) cue versus null
specificity per realization, (6) the observable endpoint. No analysis not listed here is
run on the test split.

\subsection{The decision rule}

The verdict on each model is not read off the point estimates. It is fixed by two gates and
one access criterion, all three written into the protocol before the
test split was opened, and all three computed the same way for every model, seed, intervention
and endpoint.

\textbf{Gate A, variable specificity.} For a given realization, intervention and endpoint, let
$\theta$ be the cue direction's effect and $\theta_0$ the null direction's effect on the same
items. The specificity ratio is
\[
R \;=\; \frac{\lvert \theta_0 \rvert}{\lvert \theta \rvert},
\]
the share of the cue direction's effect that a direction carrying no cue identity already
reproduces. Gate A passes when the upper bound of the 95 percent confidence interval on $R$
falls below one half,
\[
\mathrm{UCI}_{95}(R) \;<\; 0.50.
\]
The interval is a percentile bootstrap over items, 10{,}000 resamples, with the cue effect and
the null effect recomputed together on each resample so that the pairing between them is kept.
The threshold is a bound on the nuisance share, not on significance: at $R = 0.50$ half of the
apparent effect is already delivered by a direction that knows nothing about which cue was
present, and we treat that as too much to call the evidence variable specific. The bound is
taken on the upper end of the interval, so a ratio that is small but poorly resolved does not
pass.

\textbf{Gate B, practical zero.} A ratio is unstable when its denominator is near zero, which is
exactly the regime a null result produces. Gate B therefore asks the same question on an
external scale that does not shrink with the cue effect. It passes when the null direction's
effect interval lies entirely inside $\pm 0.10$ of the natural cue swap, that is when
\[
\mathrm{CI}_{95}(\theta_0) \subset [-0.10,\, +0.10]
\]
on the $\Delta$ scale of Appendix B.6, where 1 is the movement produced by actually changing the cue.
Gate B requires the null effect to lie inside the fixed practical equivalence band.
It is a check on the
comparison's validity rather than evidence for access, and it is reported whether it passes or
fails.

\textbf{The access criterion.} A model is recorded as satisfying the variable specificity
criterion only when three things hold together in the same realization: the cue direction's own effect
interval excludes zero, Gate A passes, and Gate B passes. Any one of the three failing is a
failure of the criterion. Satisfying the criterion is necessary for the causal reading of
\S2.4 and is not sufficient for it: that reading also requires comparable realized edit
magnitudes, audited in B.9, and a contrast that identifies the variable claimed, which
Section 3 states this design does not deliver. The criterion is never satisfied by a combination across seeds,
endpoints or interventions, since necessity and interchange answer different questions and
\S2.3 forbids combining them. No threshold here was chosen or moved after any
test split number was seen, and the gates are reported as they fall, including the cases where the
covariance matched comparison passes Gate A while the shuffled label comparison on the same
model fails it.

\subsection{Numerical floor}

Weights are streamed in bfloat16 and every readout is computed in float32. The smallest effect size the
protocol commits to detecting was verified on each model to sit four to five orders
of magnitude above the endpoint's numerical resolution, and a self donor intervention
(the state replaced by its own value) reads exactly zero. Both checks were passed before test split analysis
began and are not repeated per run.

\subsection{Realized edit magnitude audit}

Every direction in this section is unit normalized, but a unit direction does not fix how
much of the state it removes or replaces, since erasure takes out exactly the component of
$h$ that already lies along $v$, and that projection differs between directions even when
every direction has norm one. The audit reports the size of the edit actually realized by
each intervention, for the cue direction and for both nulls, on the same held out items used
for the estimates. The necessity edit norm is the absolute projection of the
state onto the direction,

\[
m_N \;=\; \lvert \langle h, v \rangle \rvert,
\]

and the interchange edit norm is the absolute projection of the counterfactual minus clean
state difference onto the direction,

\[
m_S \;=\; \lvert \langle h_i(X') - h_i(X),\, v \rangle \rvert.
\]

Paired null to cue ratios of these medians are reported by model, seed and intervention. The
audit is diagnostic only: it does not modify, rescale, rerun or reclassify any
reported result, and no threshold or gate is defined on it. Its numbers are read alongside the
specificity results in Section 6 to establish whether a given comparison can be attributed
to cue identity or is instead confounded by an imbalance in realized edit size.

%% file: appendix/C_results.tex
\section{Test split results in full}

This appendix carries the per model detail behind the summary in Section 6. Nothing here is a separate analysis. It is the same test split analysis, reported cell by cell.

\subsection{Transport: the donor identity decomposition in full}

\textbf{The donor identity decomposition.} (Transport estimator, authority channel, Qwen3-1.7B development split, $n = 40$.) Patching
decision circuit states from a
donor into the explanation pass first produced a readout movement of 0.194 (native donor)
and 0.189 (counterfactual donor), and the equality between the two is the finding: the
effect was the same size whichever cue the donor carried, because index wise patching had
laid cue token states onto body tokens. With a token length matched neutral prefix (``did not
conclude what the answer is'') the decomposition into generic and cue specific components is

\begin{table}[h]
\centering
\footnotesize
\caption{Donor identity decomposition of the transport readout.}
\label{tab:donor-decomposition}
\begin{tabular}{lccc}
\toprule
arm & $G$ (generic) & $C$ (cue specific) & $\Delta_{\text{ID}}$ \\
\midrule
neutralized & $-0.0051$ $[-0.050, +0.035]$ & $-0.0009$ $[-0.020, +0.011]$ & $-0.0017$ $[-0.041, +0.022]$ \\
invalid option sham control & \textbf{$-0.0534$ $[-0.102, -0.013]$} & $+0.0012$ $[-0.026, +0.022]$ & $+0.0023$ $[-0.051, +0.045]$ \\
\bottomrule
\end{tabular}
\end{table}

On the primary arm the donor identity contrast is indistinguishable from zero at an interval
37$\times$ its point estimate. The invalid option sham control shows what a generic effect looks like: an
impossible recommendation manufactures a donor effect ten times the real one with no cue
specificity at all. By the fixed development split stopping rule the transport estimator stops here.
\emph{Permitted
conclusion: a generic perturbation is not cue transport. The interval still admits a true
$C$ of about $\pm$2\% of the natural cue effect.}

\subsection{The multiclass cue direction, development and selection}

\textbf{Development and selection splits} (item level percentile bootstrap, 10,000 resamples):

\begin{table}[h]
\centering
\footnotesize
\setlength{\tabcolsep}{4pt}
\begin{tabular}{llcc}
\toprule
split & direction & necessity $N$ & interchange $S$ \\
\midrule
development, $n=35$ & cue & $-0.0019$ $[-0.0089, +0.0045]$ & $-0.0009$ $[-0.0019, +0.0001]$ \\
 & shuffle & \textbf{$-0.0094$ $[-0.0155, -0.0033]$} & $+0.0000$ $[-0.0004, +0.0005]$ \\
 & covariance matched & $+0.0014$ $[-0.0069, +0.0090]$ & $+0.0000$ $[-0.0001, +0.0002]$ \\
selection, $n=58$ & cue & $-0.0030$ $[-0.0070, +0.0012]$ & $-0.0009$ $[-0.0018, +0.0001]$ \\
 & shuffle & $-0.0005$ $[-0.0047, +0.0038]$ & $-0.0004$ $[-0.0011, +0.0002]$ \\
 & covariance matched & \textbf{$+0.0063$ $[+0.0013, +0.0115]$} & $+0.0000$ $[-0.0001, +0.0001]$ \\
\bottomrule
\end{tabular}
\caption{Necessity and interchange effects, cue direction and nulls, development and
selection splits.}
\label{tab:e5b-devselect}
\end{table}

On both splits the cue direction's intervals span zero for both claims, and on both a null
direction produces a necessity effect at least as large in magnitude with an interval that
excludes zero. Which null does so differs between splits (the shuffled identity null on
development, the covariance matched null on selection), which argues against a pathology
specific to
one null construction: the stable fact is $|\Delta_{\text{cue}}| \not> |\Delta_{\text{null}}|$.
Interchange is under 0.1\% of the natural cue swap throughout.

\textbf{Test split} ($n = 200$ requested, seeds $\{0, 1, 2\}$).

\emph{Integrity.} The same data manifest was used across all three seeds: 200/200 items
realized, 6 nonestimable, 194 per cell for every arm (cue, shuffle, covariance matched), no nonfinite or missing values, and the deterministic cue arm exactly identical across
seeds. All fixed integrity checks passed.

\emph{Representation.} $R^2_{\text{cue}}$ at layer 3 is $+0.938$ (validation of the
representation only, not a use claim).

\emph{Cue direction} (one deterministic estimate, item bootstrap):

\begin{table}[h]
\centering
\caption{Cue direction's necessity and interchange effects, test
split.}
\label{tab:e5b-test-cue}
\begin{tabular}{lcc}
\toprule
direction & necessity $N$ & interchange $S$ \\
\midrule
cue & $+0.0000$ $[-0.0029, +0.0030]$ & $-0.0002$ $[-0.0007, +0.0003]$ \\
\bottomrule
\end{tabular}
\end{table}

\emph{Null realizations}, each with its own item bootstrap:

\begin{table}[h]
\centering
\footnotesize
\caption{Null direction necessity and interchange effects by seed, test
split.}
\label{tab:e5b-test-nulls}
\begin{tabular}{llcc}
\toprule
null & seed & necessity $N$ & interchange $S$ \\
\midrule
shuffle & 0 & $-0.0022$ $[-0.0053, +0.0010]$ & $-0.0002$ $[-0.0005, +0.0001]$ \\
shuffle & 1 & $-0.0026$ $[-0.0055, +0.0005]$ & $-0.0001$ $[-0.0004, +0.0001]$ \\
shuffle & 2 & $-0.0010$ $[-0.0031, +0.0011]$ & $-0.0001$ $[-0.0003, +0.0001]$ \\
covariance matched & 0 & $+0.0052$ $[+0.0020, +0.0086]$ & $-0.0000$ $[-0.0001, +0.0001]$ \\
covariance matched & 1 & $+0.0055$ $[+0.0024, +0.0086]$ & $-0.0000$ $[-0.0001, +0.0000]$ \\
covariance matched & 2 & $+0.0030$ $[-0.0002, +0.0061]$ & $-0.0000$ $[-0.0001, +0.0000]$ \\
\midrule
\multicolumn{4}{l}{\emph{Across seeds (mean, range):}} \\
shuffle & all & $-0.0019$, $[-0.0026, -0.0010]$ & $-0.0001$, $[-0.0002, -0.0001]$ \\
covariance matched & all & $+0.0046$, $[+0.0030, +0.0055]$ & $-0.0000$, $[-0.0000, -0.0000]$ \\
\bottomrule
\end{tabular}
\end{table}

\emph{Specificity per realization}, $\Delta_{\text{cue}} - \Delta_{\text{null}}$ with its item
bootstrap:

\begin{table}[h]
\centering
\footnotesize
\caption{Cue minus null specificity per realization, test split.}
\label{tab:e5b-test-specificity}
\begin{tabular}{llcc}
\toprule
comparison & seed & necessity $N$ & interchange $S$ \\
\midrule
cue $-$ shuffle & 0 & $+0.0022$ $[-0.0012, +0.0059]$ & $+0.0000$ $[-0.0004, +0.0005]$ \\
cue $-$ shuffle & 1 & $+0.0026$ $[-0.0012, +0.0063]$ & $-0.0000$ $[-0.0005, +0.0004]$ \\
cue $-$ shuffle & 2 & $+0.0011$ $[-0.0023, +0.0045]$ & $-0.0001$ $[-0.0005, +0.0004]$ \\
cue $-$ covariance matched & 0 & \textbf{$-0.0052$ $[-0.0097, -0.0006]$} (CI excludes zero) & $-0.0002$ $[-0.0007, +0.0003]$ \\
cue $-$ covariance matched & 1 & \textbf{$-0.0055$ $[-0.0099, -0.0013]$} (CI excludes zero) & $-0.0001$ $[-0.0007, +0.0004]$ \\
cue $-$ covariance matched & 2 & $-0.0029$ $[-0.0068, +0.0009]$ & $-0.0002$ $[-0.0007, +0.0004]$ \\
\bottomrule
\end{tabular}
\end{table}

Realizations $\times$ endpoints with a cue excess reliably signed in the use direction: 0 of 12.

\textbf{(b) The substitution fails on the test split as on development and
selection.} Where a
null realization's confidence interval excludes zero, its sign is opposite the use
direction (the covariance matched null's necessity effect exceeds the cue's own, mean
$+0.0046$ across seeds vs.\ the cue's $+0.0000$, exactly the pattern already seen on select),
and no realization or endpoint gives the cue direction a reliably signed excess over its
null. Permitted conclusion: recoverability is not causal use. Per \S2.4 this asymmetry
licenses ``not established'' for the pathway this estimator tested, never ``self explanations
lack causal access.''

\textbf{Realized edit magnitude audit.} The necessity comparison above is checked against
the size of the edit each intervention actually made (Appendix B.10). On Qwen3-1.7B the erase edits
are magnitude comparable: the median null to cue ratio runs 0.79 to 1.10 across seeds for
the shuffled label and covariance matched nulls alike, so the covariance matched null's
necessity effect exceeding the cue's own is not explained by a systematically larger null
edit. The replace edits carry the opposite imbalance: both nulls receive well under half
the cue's realized interchange edit, so the near zero interchange effect on every direction
is consistent with small edits throughout rather than with a specificity finding either way.

\subsection{The observable endpoint in full}

Free form necessity and interchange, greedy 48 tokens, scored as $z(t)$ of Appendix B.6:

\begin{table}[h]
\centering
\footnotesize
\caption{Observable endpoint cell counts. For the test split the nonzero cue
interchange cell is a single item with the same value in all three seeds, and necessity
nonzero counts are given per seed (s0, s1, s2).}
\label{tab:observable-endpoint}
\begin{tabular}{l p{2.5cm} p{3.4cm} p{3.4cm}}
\toprule
split & cells exactly zero & interchange (all directions) & necessity nonzero: cue / shuffle / covariance matched \\
\midrule
development, $n=35$ & 35/36 (development check) & [0, 0] & not recorded \\
selection, $n=58$ & 338/348 & [0, 0] & 2 / 4 / 4 \\
test, $n=200$ & 1126/1123/1122 of 1164 (seeds 0, 1, 2) & [0, 0] shuffle and covariance matched, cue $+0.0067$ $[+0.0000, +0.0202]$ & 12/11/14 (seed 0), 12/14/13 (seed 1), 12/7/21 (seed 2) \\
\bottomrule
\end{tabular}
\end{table}

On select the three items that change text under a null do so under both nulls with
identical values (2.173, 0.221), so the text changes that exist are not direction specific.
At the rank-1 intervention strength the intervention rarely or never crossed a
greedy decoding argmax boundary, so no direction specific effect was detectable in
generated text. This is the resolution property anticipated in \S2.5, not an equivalence
result: the interval $[0, 0]$ is quantization, and Gate B is computed on the mechanistic
endpoint only. Intervention liveness was confirmed once by a deliberately large edit that
does change the text. The intervention strength was not raised.

On the test split, 3, 8, and 7 of the cue direction's 12 nonzero necessity cells were
also nonzero under a null, in seeds 0, 1, and 2 respectively.

\subsection{Replication tier in full}

\textbf{Qwen3-8B} (authority channel, same pipeline as Qwen3-1.7B, no redevelopment). The
development split ($n = 40$, 35 estimable) selects layer 2 by out of fold
$R^2_{\text{cue}}$ (+0.9763, versus +0.9449 at layer 3 and +0.8019 at the retired layer 12).
The selection split ($n = 60$, 58 estimable) supplies layer 2, $R^2_{\text{cue}} = +0.9711$.

\textbf{Test split.} The test sample size for this model is
$n = \lceil 429 / (1 - 5/40) \rceil = 491$, rounded to 500. This is a sample size
calculation, a different quantity from the behavioral minority arm of Section 6: the
Qwen3-8B design power shortfall there (429 required against 389 counted, power 0.860) is
not repaired by this larger sample and stands as reported in Section 6.

\emph{Integrity.} The same data manifest was used across all three seeds: 500/500 items
realized, 2 nonestimable, 498 per cell (the minimum of 429 is met), no nonfinite or missing values, and the
deterministic cue arm exactly identical across seeds. All fixed integrity checks
passed.

\emph{Representation.} $R^2_{\text{cue}}$ at layer 2 is $+0.9724$ (development +0.9763,
selection +0.9711).

\emph{Cue direction} (one deterministic estimate, item bootstrap). Necessity
$N = -0.000009$ $[-0.000145, +0.000130]$: interval spans zero. Interchange
$S = +0.000116$ $[+0.000008, +0.000224]$: interval excludes zero.

\emph{Specificity.} Of 12 realization by endpoint cells, 4 give a cue excess whose interval
excludes zero in the use direction. These do not replicate across seeds: the shuffle null
necessity excess is $-0.0001$ in seed 0 and $+0.0005$ in seed 1, opposite in sign on the
same quantity.

\emph{Gates} (\S5.5). Gate A fails in 10 of 12 realizations by endpoint, with
$R$ on necessity running 12.6 to 53.8 and upper confidence bounds as high as 104: the null
effect is one to two orders of magnitude larger than the cue's. Gate B passes in all 12,
every null interval within $\pm 0.0008$, about 125$\times$ inside the band. The
access criterion is not met.

\emph{Free form.} Exact zeros: 2975/2988, 2969/2988, 2979/2988 across seeds 0, 1, 2.

\textbf{Realized edit magnitude audit.} The necessity warning above is stronger than on
the other two models: the shuffled null received about 1.5 to 1.7 times and the
covariance matched null about 3.3 to 3.5 times the cue's realized erase edit, so the large
Gate A failures cannot be attributed to a null that was disadvantaged in edit size. On
interchange the imbalance reverses, as it does on every model: both nulls receive well
under half the cue's realized replace edit, so the statistically resolved interchange
effect that looked favorable cannot cleanly establish variable specificity either, since a
larger cue edit is itself an alternative explanation for it.

Qwen3-8B illustrates why statistical significance alone is insufficient. The cue direction
interchange effect is statistically resolved on the test split, but its magnitude
is approximately one hundredth of one percent of the natural cue swap effect, and the
target fails the fixed variable specificity gate. A statistically detectable
intervention effect does not, by itself, establish specific causal use. The
protocol anticipated this pathology: a ratio scaled specificity region collapses
when the target estimate is small, which is why Gate B exists on an external scale.

\textbf{Llama-3.2-3B} (authority channel, same pipeline as Qwen3-1.7B and Qwen3-8B, no
redevelopment). The development split ($n = 40$, 35 estimable) and the selection
split ($n = 60$, 58 estimable) both select layer 2 by out of fold $R^2_{\text{cue}}$
(development +0.9447, selection +0.9444, against the retired layer 12 at +0.5863 on
development).

\textbf{Test split.} The test sample size for this model is
$n = \lceil 296 / (1 - 5/40) \rceil = 339$, rounded to 340. This is a sample size
calculation, a different quantity from the behavioral minority arm of Section 6, where the
Llama-3.2-3B requirement of 296 is already met by the counted minority of 336 (design
power 0.940, Section 6).

\emph{Integrity.} The same data manifest was used across all three seeds: 340/340 items
realized, 5 nonestimable, 335 per cell for every arm (cue, shuffle, covariance matched), no nonfinite or
missing values, and the deterministic cue arm exactly identical across seeds. All
fixed integrity checks passed.

\emph{Representation.} $R^2_{\text{cue}}$ at layer 2 is $+0.9546$ (development +0.9447,
selection +0.9444).

\emph{Cue direction} (one deterministic estimate, item bootstrap). Necessity
$N = +0.0028$ $[+0.0022, +0.0035]$: interval excludes zero, the largest cue effect
measured anywhere in this work, about 0.28\% of the natural cue swap. Interchange
$S = -0.0000$ $[-0.0003, +0.0003]$: inert.

\emph{Null realizations}, each with its own item bootstrap. Shuffle necessity by seed:
+0.0018, +0.0022, +0.0017 (mean +0.0019, stable across seeds). Covariance matched necessity by
seed: $-0.0010$, $-0.0003$, $-0.0007$ (mean $-0.0007$). All null interchange values sit
within $\pm 0.0003$ of zero.

\emph{Specificity per realization}, $\Delta_{\text{cue}} - \Delta_{\text{null}}$, paired
by item. Cue minus covariance matched necessity is +0.0038, +0.0032, +0.0035, with the confidence
interval excluding zero in every seed. Cue minus shuffle necessity is +0.0011 (excludes
zero), +0.0007 (spans zero), and +0.0011 (excludes zero). Every interchange comparison
stays within $\pm 0.0003$ of zero. Realizations by endpoint with a cue excess reliably
signed in the use direction: 5 of 12.

\textbf{The shuffled label control.} \emph{Gates.} Gate A fails in 11 of 12 realizations by endpoint. The decisive
control is the shuffled label null. The Llama result shows why a geometry matched random
direction is not enough. The cue direction clearly beats that control, yet a direction
generated by the same fitting pipeline with cue identity scrambled reproduces most of the
target effect. On necessity the shuffled label null reproduces 61 to 76\% of the cue's
effect across the three seeds, with about 70\% a useful descriptive summary, against a
required upper confidence bound below 0.50 (the observed upper bounds run as high as
1.00). The shuffled label null comes from the same estimator at the same site under the same
normalization, and its realized edit magnitudes are comparable to the cue's, which is why it
is the stricter and more informative control here. It is not matched on the option
transition graph, as Section 5 reports, so it stress tests the inference rather than
isolating cue identity on its own. The single Gate A pass is the covariance matched necessity
comparison in seed 1.

This does not contradict the specificity table above. The paired cue minus covariance matched
difference asks whether the cue exceeds that null, and it does, with a confidence
interval excluding zero in all three seeds. Gate A asks a stricter question, a ratio of
magnitudes with its own upper confidence bound. The covariance matched necessity
comparison satisfies Gate A only in seed 1. The corresponding gate fails in
seeds 0 and 2.

\textbf{Realized edit magnitude audit.} The shuffled label comparison is
magnitude comparable: the median null to cue ratio for erase is 0.83 to 1.04 across seeds,
and the shuffled label direction nevertheless reproduces 61 to 76\% of the cue effect, so
the principal specificity failure here is not explained by systematically larger null
edits. The covariance matched comparison is not comparable in this sense: that null
received about 1.9 times the cue's realized erase magnitude, so its raw effect cannot be
read as evidence that a comparable perturbation exceeds the cue intervention. The
interchange comparison carries the reverse imbalance found on every model: both nulls
receive well under half the cue's realized interchange edit.

Gate B passes in all 12 realizations by endpoint.

\emph{Free form.} Exact zeros: 1987/2010, 1990/2010, 1988/2010 across seeds 0, 1, 2. Of
the cue direction's 10 nonzero necessity cells, 5, 5, and 3 are also nonzero under a null
in seeds 0, 1, and 2 respectively.

On Llama-3.2-3B the recovered cue identity direction is recoverable at $R^2$ 0.95, and its
erasure moves the explanation readout by about 0.28\% of the natural cue swap, the largest
such effect measured in this work. A shuffled label direction of the same construction
reproduces most of that movement. Recoverability is not causal use. The asymmetry rule of
\S2.4 applies as it does above.

\textbf{Cross model comparison.} Qwen3-1.7B: the target fails against erasure controls
whose realized edit magnitude is comparable to the cue's own. Llama-3.2-3B: a substantial
cue necessity effect survives the covariance matched, geometry only control, but the
magnitude comparable shuffled label control reproduces most of it. Qwen3-8B: a
statistically resolved but tiny interchange effect, and a realized magnitude imbalance
between the cue and its nulls makes its target versus null specificity comparison
additionally hard to interpret. Across all three models, the fixed
variable specificity criterion is not satisfied. We do not attribute these differences to
scale, architecture, or model family.

\textbf{The audit cuts both ways.} In two places a larger null edit could explain a
negative looking comparison: the covariance matched necessity null on Llama-3.2-3B and both
necessity nulls on Qwen3-8B were each given a larger realized edit than the cue, which is a
plausible account for their comparatively large null effects. In two other places a smaller
null edit could explain a favorable looking cue comparison: every interchange null on all
three models received well under half the cue's realized edit, which is a plausible account
for the near zero interchange effect from every null, favorable looking cue interchange
comparisons included. The audit is read as a validity check on both directions of
interpretation, not as protection for the paper's own reading of these results.

Scale does not account for the pattern: the ordering across 1.7B, 3B, and 8B is
nonmonotonic, with the largest model showing the smallest effect (roughly 0, 3e-3, and
1e-4 of the natural cue swap respectively). Family is not claimed either, since the set
contains two Qwen models and one Llama model, and family is confounded with tokenizer,
training data, instruct tuning, and architecture. Descriptively, and with no cause
attributed, the two Qwen models behave alike (near zero to very small effects) while
Llama differs (a larger, more reproducible effect that still fails the shuffled label
control).
The cross model summary is Table~\ref{tab:cross-model} in Section 6 and is not
repeated here.

The gates are computed identically for all three models. On Qwen3-1.7B, Gate A fails in all 12
realizations by endpoint, with the ratio on necessity running from 23.2 to 124.4 and its upper
confidence bounds from 11.8 to 54.3. The upper bound sits below the point estimate, which is
not an error but the expected behavior of a ratio whose denominator is near zero: the cue's
necessity effect here is $+0.000044$, so resampled values of the denominator are almost always
larger than the point value and the resampled ratios are correspondingly smaller. This is
precisely the regime in which \S5.5 directs that the ratio be reported and not interpreted
substantively, since the operative fact is that there is no resolved target effect rather than
that a null reproduced one. Gate B passes in all 12.

The replication set is complete: all three models (Qwen3-1.7B,
Qwen3-8B, Llama-3.2-3B) have test split analyses. No further estimator is added.
The one further model is the additional model reported in C.5 below, which is
not part of the main claim.

Reported per model, never pooled.

\subsection{The fourth model, Llama-3.2-1B, in full}

Run after the other three, with the site rule, the nulls, the endpoints, the gates, the
thresholds, the seeds and the sample size all fixed before its test split was touched, and its
layer chosen on its own development split rather than inherited from the 3B.

\emph{Behavioral gate}, full pool, 2,116 items, manifest identical to the other models.
Flip 29.7 percentage points against the 3B's 63.8, adjudicated mention 1.37\% (29 of 2116),
usable 1096, followed 1111, resisted 1005, all five criteria pass.

\emph{Site.} Layer 2, selected on the 1B's own development split by the out of
fold criterion and not inherited from the 3B.

\emph{Integrity.} $n = 340$ requested and realized, 5 non-estimable, 335 per cell for every
arm, one manifest across seeds, deterministic cue arm identical across seeds, no non-finite or
missing values.

\emph{Representation.} $R^2_{\text{cue}} = +0.9413$ at layer 2, against +0.9340 on development
and +0.9417 on selection.

\emph{Effects.} Cue necessity $-0.000567$ $[-0.001271, +0.000118]$, spanning zero. Cue
interchange $+0.000467$ $[+0.000198, +0.000750]$, excluding zero. The excess over both nulls
excludes zero on interchange in all three seeds, at $+0.00036$ to $+0.00039$ against the
shuffled label null and $+0.00045$ to $+0.00048$ against the covariance matched null. Necessity
gives nothing against either null in any seed.

\emph{Selection did not persist.} On the selection split the cue necessity effect was
$+0.0018$ $[+0.0003, +0.0035]$, an interval excluding zero with neither null resolved, which on
the test split became $-0.0006$ spanning zero.

\emph{Gates.} Gate A passes in 6 of 12 realizations by endpoint, all six of them interchange,
with ratios of 0.16 to 0.23 against the shuffled label null and 0.01 to 0.04 against the
covariance matched null. Gate B passes in all 12. The access criterion is therefore
satisfied on interchange in all three seeds against both nulls.

\emph{Realized edit magnitude.} Erasure is comparable, with paired null to cue medians of 0.88
to 1.23 for the shuffled label null and 0.83 to 0.97 for the covariance matched null.
Replacement is not, at 0.355 to 0.385 and 0.077 to 0.087, so the cue direction receives roughly
three and twelve times the edit its controls receive on the endpoint that passed.

\emph{Observable endpoint.} Exact zeros 1988, 1980 and 1980 of 2010 across seeds 0, 1 and 2.

\emph{What it supports.} The criterion is met. Its causal reading has two live alternative explanations that this design cannot
separate from cue identity: the unequal realized replacement magnitudes above, and the fact
that the cue contrast changes cue identity and cue answer agreement together. Nothing
was rerun, rescaled or reclassified to address either.

%% file: appendix/D_limitations.tex
\section{Limitations in full}

\textbf{Artificial cue.} The cue used throughout is an injected authority statement naming a
specific wrong option: a strong, known cause chosen so its causal role in the decision
could be established by construction and measurement, the first rung of the ladder in Section 1. This buys certainty about
what $C$ is and that it drives $y$, but it is not a naturally occurring bias. A bias the
model acquired during training, rather than supplied in context at inference time, may be
weaker, distributed across more components, represented in a form that does not sit close to
a token span, or absent from the residual stream in any literal sense. Nothing here licenses
extrapolating the specificity failures to that different, harder case.

\textbf{Model set.} Four models are covered, drawn from two families, two Qwen models and two
Llama models, and every result
is reported per model (flip rate, mention rate, design power), never pooled into a single
number. This is deliberate: pooling across families would treat family identity as a
nuisance variable, which the conditional verbalization result (Section 6) suggests it is
not. It also means each estimate rests on a single model rather than a family average, and
the Qwen3-8B arm's design power of 0.860 is below the 0.90 target fixed in the
protocol. That shortfall is reported as measured, not adjusted for. The 8B results
carry correspondingly less power to detect a real effect than the other two arms.

\textbf{Intervention family.} Every intervention in this paper is either a rank-1 edit at a single
residual stream position or a whole component patch between matched forward
passes. Neither family covers multi position interventions, nonlinear edits, or
interventions on attention patterns rather than on the residual stream. A pathway that
requires coordinated changes across positions, or that is not linearly represented at the
tested site, is invisible to every estimator in Section 5 by construction, independent of
whether it exists.

\textbf{Early layer selection.} The representation estimator's site, layer 3, was chosen by
out of fold $R^2_{\text{cue}}$ on a development split: the layer where the cue is most
literally recoverable. Section 8 argues this is expected to be an early, token proximal
layer regardless of where the cue is causally used, because the selection criterion cannot
distinguish encoded from used. A causally motivated site selection rule (e.g., choosing the
layer by a pilot intervention's effect size) would face the opposite problem the
protocol was built to prevent: selecting a site by an intervention's outcome biases
the specificity test that follows. No such rule was specified for this paper, and none is
retrofitted after seeing the layer-3 null result. A causally motivated criterion, if one is
to be used, has to be fixed before any intervention outcome is examined: work for a future
study, not a correction applied here.

\textbf{Greedy free form resolution floor.} The primary observable endpoint (greedy decoded
free form text, scored by log probability under the counterfactual context) is quantized at
the intervention strength: most cells are exactly zero because no intervention
crossed an argmax boundary (Section 6). Liveness was confirmed directly, by showing that a
substantially larger edit at the same site does change the generated text, but this
establishes only that the intervention mechanism works, not that the intervention strength has power on this
endpoint. That sensitivity question is not established, and the strength was deliberately
not raised to chase a nonzero cell, since doing so would be a search over strengths rather
than a measurement at a single fixed strength.

\textbf{Asymmetry of the conclusion.} By the rule fixed in Section 2.4, failure of these three
estimators to produce cue specific evidence surviving their controls does not establish that
the explanation computation lacks causal access to the cue. It establishes that these
particular interventions, at these sites, on these endpoints, at these strengths, did not
produce it. Pathways these estimators cannot rule out include: cue information carried in a
subspace other than the one tested, a nonlinear encoding no linear or component level
intervention would disturb, an explanation specific recoding of the cue unlike its
decision pass representation, and components entirely outside the tested attribution and
transport sets.

\textbf{Single cue channel in the test arm.} The main analysis is run on
one cue channel (authority) chosen because it produced the strongest and cleanest
behavioral effect during development (Section 6). Two other channels (explicit, positional)
were characterized earlier in this project and are not part of the main claim. The
extent to which the specificity falsifications generalize across cue channels is a
development stage observation only, not a result.

\textbf{Mention measurement.} Section 3 gives the precision and recall adjudication in full,
including the 8 false negatives found across 8,464 explanations and the finding that 7 of them
were censored by the generation cap rather than missed by vocabulary. Two consequences belong
here. The corrected rates remain estimates rather than exact counts, since a screen can only
bound what it did not find. And mention is, by construction, a posttreatment variable
(observed only after the cue has already been shown to move the decision), and none of the
causal access analyses condition on it. It is reported descriptively, never used as a
covariate or a filter anywhere in the ladder.

\textbf{Explaining a forced answer is not the same as explaining a chosen one.} The answer being
explained is held fixed at the cued option in every condition, so an intervention can change
only the explanation. On items where the model followed the cue, the explanation prompt asks
about an answer the model did in fact give. However, on items where the model resisted the
cue, the same prompt asks it to justify an answer it did not choose. Those passes are
forced answer justifications rather than reports on a preceding decision. The mechanistic
analyses pool both kinds, because conditioning on which arm an item fell into would condition
on a posttreatment variable. The two are not the same object, and any reading of these
results as being about self report of a real choice applies cleanly only to the followed arm.

\textbf{The counterfactual moves two things at once.} Switching the cue from the named option to
the counterfactual option changes which option the cue names, but it also changes whether the
cue agrees with the answer being explained, since the answer stays fixed. Under the original
cue the stated expert opinion matches that answer, while under the counterfactual it
contradicts it. The contrast therefore carries cue identity and cue answer agreement
together, and this design cannot separate them. A design that varied identity while holding
agreement fixed would need a different construction and is not attempted here.

%% file: appendix/E_estimators.tex
\section{The three estimators in detail}

Section 2 argued that three kinds of mechanistic evidence are routinely offered for a
causal access claim, that each rests on an inference from a signature compatible with use
to use itself, and that each inference has a distinct nuisance alternative that a
generic version of the same statistic would also produce. This section applies that
framework to three increasingly targeted estimators, run in the order the framework
motivates: attribution, then transport, then recoverable representation. We test three
increasingly strong forms of evidence that could support a causal access claim. Each
requires a different specificity control. In our setting each apparent positive fails
that control. The three subsections share one template (apparent positive evidence, the
falsification control, what the control found in development, and the conclusion the
evidence permits) because the failures are not three
unconnected null results but three instances of the same gap, exposed by three different
controls because the three estimators are exposed to three different nuisance
alternatives.

\subsection{Component attribution}

\textbf{Evidence.} Component attribution is the most direct form of
mechanistic evidence: ablate or patch one component of a pass at a time, score the
change in a readout, and ask whether a small set of components accounts for most of the
effect. On the decision pass (v1 of this project, the explicit cue channel, Qwen3-1.7B,
$n=35$), this is exactly what happens, and it is genuine: the top 24 of 476 components
(5\%) recover 91\% of the full cue manipulation effect at 2.85x a layer and kind matched
random baseline. This is the decision side reference localization the rest of the framework is
measured against: $D$ demonstrably contains a concentrated pathway from the cue to the
answer. On the explanation pass, the same style of attribution, read through a
single position KL divergence between the clean and cue swapped next token
distributions, shows the same qualitative shape: the top 24 components recover the
effect at 2.42x random. Read on its own, that number says component attribution
localizes on the pass the paper actually cares about, at a ratio close to the decision
pass's.

\textbf{Control.} Section 2.2's nuisance alternative for attribution is
that the readout is sensitive to \emph{generic} component importance rather than to
cue specific content: any component that matters for producing fluent text at all will
register on a readout that depends on that text, and on a weak signal the generic term
dominates. The control is therefore not a random component baseline but a
layer and kind matched one, together with a power gate evaluated on an
independent subset: effect over sd $\geq 2$ and top over random $\geq 2$, fixed before
the gate was run. A second, sharper check rescores the decision pass's own construct
(the pathway already known to be real) as a full sequence sum instead of a single
divergent position score, to ask whether the localization survives a change in \emph{how much
of the sequence} the attribution is asked to explain.

\textbf{Development result.} The power gate fails on the explanation pass
readout: effect over sd is 0.57 and top over random is 1.85, both below the
threshold of 2. The 2.42 seen at $k=24$ did not replicate on the independent subset the
gate is computed on. More telling is the second check, because it is applied to the
decision pass's \emph{own} pathway (the one already validated at 2.85x) and not to a new
readout under suspicion: rescored as a full sequence sum, the same construct collapses
from 2.85x to 1.28x, with a single random component reproducing 104\% of the full effect
at $k=1$. Full sequence attribution does not localize even on a circuit known to exist.
That result reframes what the 2.42 on the explanation pass could have meant: a
single position, high mass readout is the only form that has ever localized in this
project, and the explanation pass's whole footprint is small enough (a 0.074-nat KL
between clean and counterfactual) that the generic term is positioned to dominate it.

\textbf{Interpretation.} Concentration is not access. A sparse, high scoring
set of components is what a readout sensitive to generic importance produces whether or
not any of the components in it carries cue identity, and the power gate
(built precisely to distinguish the two) is not passed. The design lesson carried
forward to the transport and representation estimators below is the same one stated in Section 2.3: a metric localizes
only if it is single position and high mass, and on a signal this weak, a control that
does not match generic importance will manufacture an apparent localization on its own.

\subsection{Whole component transport}

\textbf{Evidence.} A stronger form of evidence than attribution within a
single pass is transport across passes: take the decision circuit components'
activations under the cue and patch them into the explanation pass, and ask whether the
explanation readout moves. On the authority channel (development split, $n=40$),
patching a donor run's decision circuit states into the explanation pass produces a
readout of 0.194 under the native cue donor and 0.189 under a counterfactual cue donor.
Read naively, an effect of this size (a fifth of the natural cue swap contrast) looks
like a substantial transport effect: the decision pass's representation was carried
into the explanation pass and moved it.

\textbf{Control.} The nuisance alternative here is that patching \emph{any}
foreign activation into the explanation pass moves the readout, independent of what
identity it carries: through positional misalignment, through a donor that simply
differs from the target in magnitude or fluency relevant content, or both. The control
decomposes the effect by donor identity: $\Delta_{\text{ID}} = \Delta_{X'} - \Delta_X$,
split into a generic term $G = (\Delta_X + \Delta_{X'})/2$ and a cue specific term
$C = (\Delta_{X'} - \Delta_X)/2$. Only $C$ bears on transport. $G$ is what any donor of
that magnitude and geometry would produce regardless of which cue it carries. The
control further requires a \emph{true} neutral donor (a token length matched prefix that
draws no conclusion at all, ``did not conclude what the answer is'', rather than an
offset aligned donor, which was rejected because RoPE makes positionally shifted states
noninterchangeable) and demotes an invalid label donor, ``(Z)'', naming an option that
does not exist, to a sham condition rather than a primary one.

\textbf{Development result.} The first run's 0.194/0.189 was exposed by
exactly the donor identity contrast the control specifies: the effect was the same size
whichever cue the donor carried, which is the signature of index wise patching of
cue token states onto body token positions, not of cue identity transport. On the
corrected, true neutral donor, the decomposition gives $G = -0.0051$
$[-0.050, +0.035]$, $C = -0.0009$ $[-0.020, +0.011]$, and
$\Delta_{\text{ID}} = -0.0017$ $[-0.041, +0.022]$: an interval roughly 37 times the
point estimate, spanning zero throughout. The invalid option sham control is the sharper
demonstration of why the demotion mattered: it manufactures $G = -0.0534$
$[-0.102, -0.013]$, a generic effect whose interval excludes zero and is roughly ten
times the size of the true neutral donor's, while its cue specific term, $C = +0.0012$,
is indistinguishable from zero. Telling the model an expert recommended an impossible
option changes the computation enough to produce a large donor effect with no cue
specificity at all. Had the sham been left as the primary condition, this would have
been read as a confident transport result. A further check (a paraphrased authority
donor naming the same option as the unmodified sentence) was meant to serve as a
semantic equivalence control and failed at that role instead: it shifted the raw
endpoint by about 4.5 times the minimum detectable effect, so it was kept
only as a held out environment, not as evidence of anything. By the fixed
development stopping rule, the transport estimator halted at this point.

\textbf{Interpretation.} A generic perturbation is not cue transport. The
0.19 that looked like a transport effect was a positional artifact, and once donor
identity and a true neutral condition are in place, the cue specific term is
indistinguishable from zero with a confidence interval wide enough to admit a true
effect of only a few percent of the natural cue swap magnitude, narrow enough to rule
out a large transport effect, wide enough that this is a report of ``not established,''
not of ``transport is absent.''

\subsection{Recoverable cue representation}

\textbf{Evidence.} The most targeted estimator asks not whether a
component matters or whether an activation transports, but whether the cue's identity
is itself linearly recoverable at a site in the explanation pass, and whether
intervening along that direction moves the explanation causally. The first attempt
(the single axis estimator), a single global direction $v = \operatorname{norm}(\sum_i d_i)$ at a
coherence selected layer, is preserved as a documented but failed construction: it did
not survive its own out of fold sign check on the development split \emph{before any
intervention was run}: held out projection $-0.044$, one item carrying 84\% of the
mean, and 14 distinct $(X, X')$ option pairs among 20 items, with unit normalizing the
contrasts making the instability worse (coherence 0.485 to 0.188). The diagnosis is
itself a representational finding: cue identity names one of up to seven options, so it
is a multiclass variable, and a single binary axis cannot encode it. The amendment to a
multiclass estimator (summarized here: the full construction, its layer
selection, interventions, nulls and endpoints are in Section 5) was made with no
causal result in hand, as the framework's fixed stopping rule requires. It models the cue
contrast as an item fixed effect plus per option effects, $\hat B = A^{+} D$,
producing a pair conditional direction for each held out item, with the layer chosen by
out of fold reconstruction ($R^2_{\text{cue}}$) on the development split alone (layer
3, the discarded single axis estimator's layer 12 scores 0.004 by this criterion). On this
construction the representational check that the single axis version failed, the multiclass one passes decisively:
$R^2_{\text{cue}} = +0.913$ on development and $+0.939$ on the selection split, both out
of fold: the cue's identity is recoverable from the explanation pass activations to a
high degree.

\textbf{Control.} The nuisance alternative for recoverability is that a
representation can be decodable and causally inert: reconstruction criteria select the
site where a variable is most \emph{literally} encoded, which for an in context cue is close
to its tokens and early in the network, and nothing in that criterion distinguishes
encoding from use. The control runs necessity (erase the component along the direction)
and interchange (replace it with the value observed under the counterfactual cue)
through the identical pipeline for the cue direction and for two null directions
constructed with the same estimation and intervention procedure (normalization, both
interventions) so that any effect a null produces is attributable to the pipeline, not
to cue identity: a shuffled identity null, which refits the same option effect model
after permuting option labels, and a covariance matched null, a random direction drawn
from the empirical activation covariance. The specificity claim is the excess of the
cue direction's effect over each null's, read separately for necessity and interchange.

\textbf{Development result.} On development ($n=35$, after 5 nonestimable
contrasts were dropped), the cue direction's necessity effect is $-0.0019$
$[-0.0089, +0.0045]$, a confidence interval spanning zero, while the shuffled identity
null's necessity effect is $-0.0094$ $[-0.0155, -0.0033]$, five times larger in
magnitude and reliably signed. A null direction, carrying no cue identity by
construction, produces a larger causal effect than the real one. On the selection
split ($n=58$), the same comparison shifts which null is the offending one: cue necessity
is $-0.0030$ $[-0.0070, +0.0012]$ against a covariance matched null of $+0.0063$
$[+0.0013, +0.0115]$, while the shuffled null on this split is small,
$-0.0005$ $[-0.0047, +0.0038]$. Interchange effects are under 0.1\% of the natural cue
swap on both splits. The stable fact across the two splits is that the cue direction's
effect is not reliably larger in magnitude than at least one null's. That the null
which exceeds it changes identity from shuffle (development) to covariance matched
(selection) argues that this is not a pathology of one null's construction, but a
property of the direction itself at this site. The test on the held out
test split, read across three seeds, is the last reading of this
comparison: the substitution fails on the test split as on development and
selection: cue
necessity is $+0.0000$ $[-0.0029, +0.0030]$ against a covariance matched null's $+0.0046$
mean across seeds, and 0 of 12 realizations $\times$ endpoints give the cue direction a
reliably signed excess over its null (Section 6).

\textbf{Interpretation.} On development and selection, recoverability is
not causal use. A direction that reconstructs the cue's identity out of fold at
$R^2$ above 0.9 need not, when intervened on, move the explanation pass by more than a
null direction carrying no cue identity, and here it does not. Per the framework's
fixed stopping rule, no further estimator (a probe, PCA, LDA, an SAE feature, CCA, or a
raised intervention strength) is fitted at this or any other site on the strength of
this result.

\subsection*{Why three different controls}

The three controls are different because the three nuisance alternatives are different,
and running one estimator's control on another's evidence would not have caught its
defect: matched importance random components say nothing about donor identity
confounds, and a donor identity contrast says nothing about whether a decodable
direction is causally inert. Each failure here was found by a control specified before
the corresponding data was seen (the power gate before the KL readout was scored on an
independent subset, the donor identity decomposition before the neutral donor run, the
shuffled and covariance matched nulls before development data touched the multiclass
direction), and none was found by a post hoc argument about what the number ``really''
meant. That is the sense in which the three results above are not three failed attempts
at the same thing, but three applications of one framework to three structurally
distinct forms of evidence, each of which turned out, in this setting, to require the
control it was given.